\documentclass{article}

\usepackage[preprint]{corl_2025} 

\usepackage{amssymb}

\usepackage{graphicx}
\usepackage[T1]{fontenc}
\usepackage{epsfig}
\usepackage{amsmath}
\usepackage{amssymb}
\usepackage{color, soul, colortbl}
\usepackage{mathtools}
\usepackage{booktabs}
\usepackage{times}
\usepackage{microtype}
\usepackage[labelfont=bf, font=footnotesize]{caption}
\usepackage[labelfont=bf, font=footnotesize]{subcaption}
\usepackage[utf8]{inputenc}
\usepackage{float}
\usepackage{inconsolata}
\usepackage{multirow}
\usepackage{placeins}
\usepackage{stfloats}
\usepackage{enumitem}
\usepackage{tabularx}
\usepackage{xstring}
\usepackage{multirow}
\usepackage{xspace}
\usepackage{url}
\usepackage{hyperref}
\usepackage{xcolor}
\usepackage{duckuments}
\usepackage{svg}
\usepackage[hang,flushmargin]{footmisc}
\usepackage{url}
\usepackage{tabularray}
\usepackage{wrapfig}
\usepackage{epigraph}

\makeatletter
\patchcmd{\epigraph}{\@epitext{#1}}{\footnotesize\itshape\@epitext{#1}}{}{}
\makeatother

% ...

\newcommand{\ssim}{{\sim}}

\newcommand{\model}{\texttt{Sparsh-X}\xspace}

\newcommand{\R}[1]{{%
    \textbf{%
        \ifstrequal{#1}{1}{\textcolor{red}{R#1}}{%
        \ifstrequal{#1}{2}{\textcolor{blue}{R#1}}{%
        \ifstrequal{#1}{3}{\textcolor{magenta}{R#1}}{%
        \ifstrequal{#1}{4}{\textcolor{teal}{R#1}}{%
                           \textcolor{cyan}{R#1}%
        }}}}%
    }%
}}

\definecolor{tabfirst}{rgb}{0.7, 1.0, 0.7} 
\definecolor{tabsecond}{rgb}{1, 1, 0.7} 
\definecolor{tabthird}{rgb}{1, 0.85, 0.7} 

\newcommand{\raisemath}[1]{\mathpalette{\raisemath{#1}}}

\newcommand{\av}{\mathbf{a}}

\newcommand{\Tv}{\mathbf{T}}

\definecolor{RebuttalColor}{RGB}{0,0,0}

\title{Tactile Beyond Pixels: Multisensory Touch Representations for Robot Manipulation}

\author{
Carolina Higuera$^{1,2*}$,
Akash Sharma$^{1,3*}$, 
Taosha Fan$^{1*}$,
Chaithanya Krishna Bodduluri$^{1}$, \\ \bf
Byron Boots$^{2}$, 
Michael Kaess$^{3}$, 
Mike Lambeta$^{1}$, 
Tingfan Wu$^{1}$, 
Zixi Liu$^{1}$, \\ \bf
Francois Robert Hogan$^{1+}$,
Mustafa Mukadam$^{1+}$\\[3mm]
$^{1}$FAIR at Meta,
$^{2}$University of Washington,
$^{3}$Carnegie Mellon University, \\
$^{*}$Equal contribution
$^{+}$Equal Advising
}

\begin{document}
\maketitle

\begin{figure}[htbp] 
    \vspace{-10mm}
    \centering
    \includegraphics[width=0.85\linewidth]{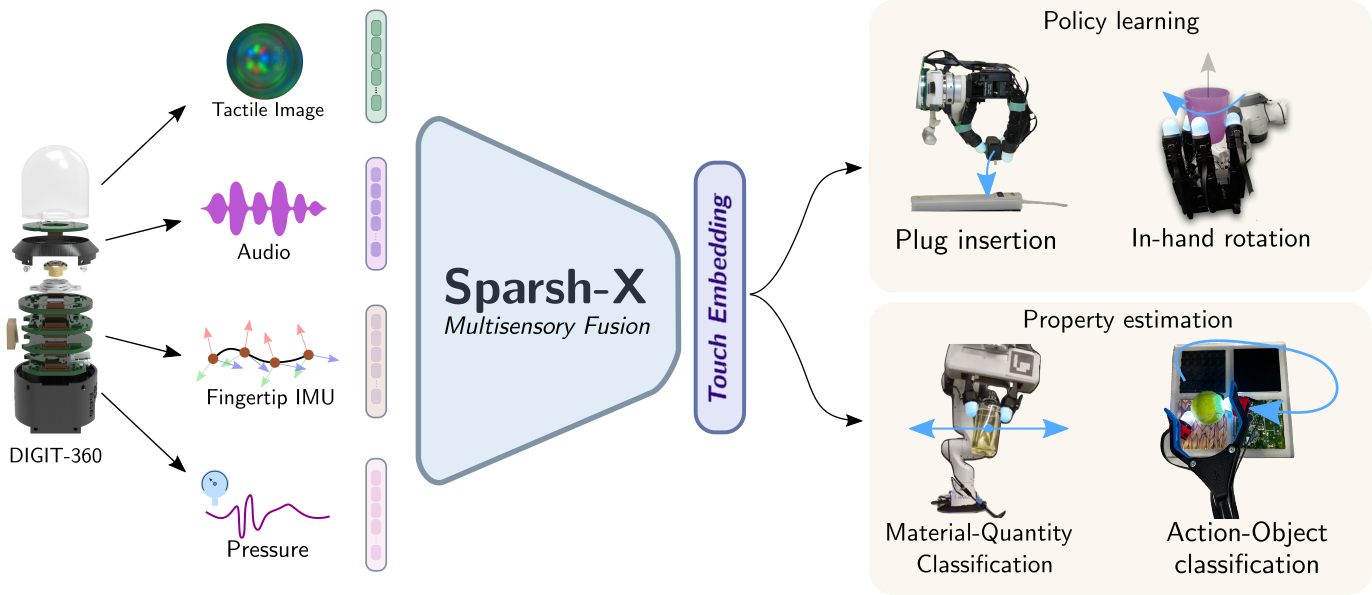}
    \vspace{-2mm}
    \caption{
        \textbf{\model, Multisensory Touch Fusion Transformer for General-Purpose Representations.}  Touch in robotics can be sensed through multiple modalities, including tactile images, vibrations, motion, and pressure. \model is a transformer-based backbone that fuses these modalities from the Digit~360 sensor. We show its versatility across diverse downstream tasks: manipulation via imitation learning (plug insertion), tactile adaptation (in-hand object rotation), and benchmark tasks to probe the understanding of physical properties.
    }
    \label{fig:teaser}
    \vspace{-5mm}
\end{figure}
\begin{abstract}
We present \model, the first multisensory touch representations across four tactile modalities: image, audio, motion, and pressure.  Trained on $\sim$1M contact-rich interactions collected with the Digit 360 sensor, \model captures  complementary touch signals at diverse temporal and spatial scales. By leveraging self-supervised learning, \model fuses these modalities into a unified representation that captures physical properties useful for robot manipulation tasks.  We study how to effectively integrate real-world touch representations for both imitation learning and tactile adaptation of sim-trained policies, showing that \model boosts policy success rates by 63\% over an end-to-end model using tactile images and improves robustness by 90\% in recovering object states from touch.  Finally, we benchmark \model's ability to make inferences about physical properties, such as object-action identification, material-quantity estimation, and force estimation. \model improves accuracy in characterizing physical properties by 48\% compared to end-to-end approaches, 
demonstrating the advantages of multisensory pretraining for capturing features essential for dexterous manipulation.
\vspace{-5mm}

\end{abstract}

\section{Introduction}
\vspace{-2mm}
Touch is a rich and multifaceted sense that plays a central role in human dexterity. Humans fluidly adapt their interactions to the physical properties of objects by integrating a wide spectrum of touch signals that include skin deformation, vibrations, motion, and pressure. This multisensory feedback enables us to distinguish between a plastic and paper cup, twirl a pen between fingers with ease, and manipulate tools under severe visual occlusion. Leveraging the multisensory nature of touch is desirable for robust, fine-grained robot manipulation.
Touch is a rich and multifaceted sense that plays a central role in human dexterity. Humans fluidly adapt their interactions to the physical properties of objects by integrating a wide spectrum of touch signals that include skin deformation, vibrations, motion, and pressure. This multisensory feedback enables us to distinguish between a plastic and paper cup, twirl a pen between fingers with ease, and manipulate tools under severe visual occlusion. Leveraging the multisensory nature of touch is desirable for robust, fine-grained robot manipulation.

Despite its importance, multisensory touch remains significantly underutilized in robotics. Most approaches rely on unimodal tactile sensing, such as GelSight-like sensors~\cite{gelsight, digit, lepora2021soft}, due to standardized hardware availability. However, advances like Digit~360~\cite{lambeta2024d360} now enable capturing high-resolution images, vibrations, motion, and pressure in a compact form, making multisensory touch accessible. Prior work on using tactile modalities independently shows promise~\cite{yu2024mimictouch, mejia2024hearing}, but a unified, scalable, and easily integrable method to take advantage of these modalities is still lacking. Representation learning offers a viable solution to integrating heterogeneous sensory inputs from sensors like Digit~360 by fusing complementary contact information from all modalities into a shared latent space. In addition, as has been shown for vision-based tactile sensors~\cite{xu2025unit, zhao2024transferable, higuera2024sparsh, gupta2025sensor} representation learning allows a downstream task training to be more data-efficient and robust to noise or irrelevant variations~\cite{oquab2023dinov2, bardes2023vjepa, hendrycks2019using, karnan2023sterling}.

This paper introduces \model, the first self-supervised backbone for multisensory touch representation learning across four key tactile modalities: image, audio, motion, and pressure. Trained on unlabeled data from diverse manipulation behaviors, such as sliding, tapping, rotating, picking up, and dropping, \model learns to fuse heterogeneous tactile signals into compact and expressive contact embeddings. Beyond \model, this unlabeled dataset also facilitates research in representation learning and benchmarking. Through supervised tasks, we show that \model captures a wide range of physical properties, including object-level characteristics (e.g., type and mass), static contact properties (e.g., force and material), and dynamic interaction cues (e.g., motion and impact). Representations that encode these physical properties at the fingertip level are especially valuable for dexterous manipulation, as they enable feedback of object and contact state directly in latent space. 

Ultimately, tactile signals are valuable only when effectively used in policy learning. This remains a challenge, particularly in reinforcement learning due to the \textit{sim-to-real} gap~\cite{lin2022tactile}. We demonstrate that \model can be effectively applied in policy training via two examples: (1) imitation learning, and (2) tactile adaptation of policies trained in simulation with privileged access to contact information. Experiments across manipulation tasks, including insertion and in-hand rotation, demonstrate that integrating \model leads to significantly improved real-world performance over end-to-end tactile image baselines. By unifying multisensory touch in a shared latent space, \model takes a step toward foundation models for touch, enabling scalable and data-efficient learning for fine-grained robotic manipulation.  Our key contributions are:

\begin{enumerate}[itemsep=-1.0pt,topsep=-2pt,leftmargin=6mm] 
    \item \model, the first unified backbone for multisensory touch: fusing image, audio, motion, and pressure signals into a general-purpose representation. Trained on $\ssim1$M unlabeled samples from Digit~360, \model enables scalable and transferable touch perception. 
    \item The first Digit~360 dataset curated for benchmarking multisensory touch representations, allowing interpretability analysis in terms of contact dynamics and physical properties of the object. 
    \item An empirical demonstration of \model enhancing real-world policy learning performance and robustness enabled by \emph{tactile adaptation} for fine-grained manipulation skills like insertion and in-hand rotation.
\end{enumerate}

\vspace{-1mm}

\section{Related Work}

\vspace{-3mm}
Vision-based tactile sensors~\cite{gelsight, digit, lepora2021soft, 8593661} have been widely used in contact-rich tasks, including material and volume prediction~\cite{8461164, 10341880}, shape inference~\cite{suresh2024neuralfeels, gao2024exploiting}, localization~\cite{suresh2023midastouch, huang2024normalflow}, insertion~\cite{dong2021tactile}, and contour-following~\cite{aquilina2024tactile, xue2025reactive} among others. While tactile sensing has been primarily vision-based, other modalities such as audio have also been used independently for capturing object properties~\cite{gandhi2020swoosh, clarke2022diffimpact} and dynamic behaviors~\cite{thankaraj2023sounds}, though audio alone may be insufficient for perceiving continuous interactions such as forces, deformations, and motion. Subsequently, prior work has also explored audio-visual learning~\cite{mejia2024hearing, liu2024maniwav} as a natural extension.  While audio and vision modalities augment the tactile state complimentarily, additional modalities such as fingertip motion and accumulated pressure can provide additional information for detecting shear forces, recognizing object properties, and predicting object slip and pose changes.

Self-supervised learning (SSL) has been effective for developing tactile image representations~\cite{feng2025anytouch, gupta2025sensor, xu2025unit, higuera2024sparsh, zhao2024transferable}, allowing better performance and data-efficiency in downstream tasks. Other approaches that use additional tactile modalities such as audio, often rely on using task-specific data to fine-tune pre-trained encoders (e.g. AST~\cite{gong2021ast} and BYOL-A~\cite{9944865} for audio) or joint audio-visual encoders~\cite{morgado2021audio}. MULSA~\cite{pmlr-v205-li23c} introduced multimodal transformers integrating vision, tactile image, and audio from contact microphones by treating all signals as RGB images, but suffers from quadratic complexity of pairwise attention since it simply concatenates the tokens from all modalities. MimicTouch~\cite{yu2024mimictouch} proposed unimodal SSL for tactile images and audio separately, without explicit cross-modal fusion. In contrast, we propose \model, a multisensory framework that fuses image, audio, motion, and pressure signals via bottleneck self-attention~\cite{NEURIPS2021_76ba9f56} into a shared latent space. This enables to capture contact properties for downstream tasks, while also reducing computational complexity compared to vanilla transformer-based concatenation strategies.

\vspace{-2mm}
\paragraph{The Digit 360 sensor.}
Digit~360~\cite{lambeta2024d360} offers multisensory touch sensing in a compact fingertip form factor, making it well-suited for dexterous robotic hardware. Inside a dome-shaped elastomer, it integrates a hyper-fisheye camera, contact microphones, IMU, static pressure sensors, and more. While it marks a significant step toward standardizing touch sensing, its unique form factor introduces new challenges. For example, the soft hemispherical dome deforms and moves  upon contact, complicating shear force estimation. Additionally, the use of a hyper-fisheye lens and directional lighting limits the applicability of photometric surface reconstruction methods like Poisson integration~\cite{gelsight}. Despite these limitations, learning touch representations through scale pretraining offers a promising path to overcome such challenges.

\vspace{-2mm}\section{Learning Multisensory Touch Representations}
\label{sec:sparsh_d360_model}

\vspace{-2mm}
Fusing tactile modalities is crucial to discovering correlation between modes. For instance high-frequency audio and tactile image can both indicate making/breaking contact. However, traditional tactile sensing work has largely relied only on unimodal approaches, i.e., on tactile images of elastomer deformations~\cite{gelsight, digit}. In cases, where additional modalities are considered~\cite{yu2024mimictouch, pmlr-v205-li23c}, they have been treated independently. 
In this section, we introduce \model, a backbone for general multimodal touch representations for the Digit~360~\cite{lambeta2024d360} sensor. Our model integrates four tactile modalities: image, audio, accelerometer, and pressure. Through self-supervision on~$\ssim1$M unlabeled contact interactions, \model compresses contact information into a unified multimodal representation.

\begin{figure}
    \centering
    \includegraphics[width=\linewidth]{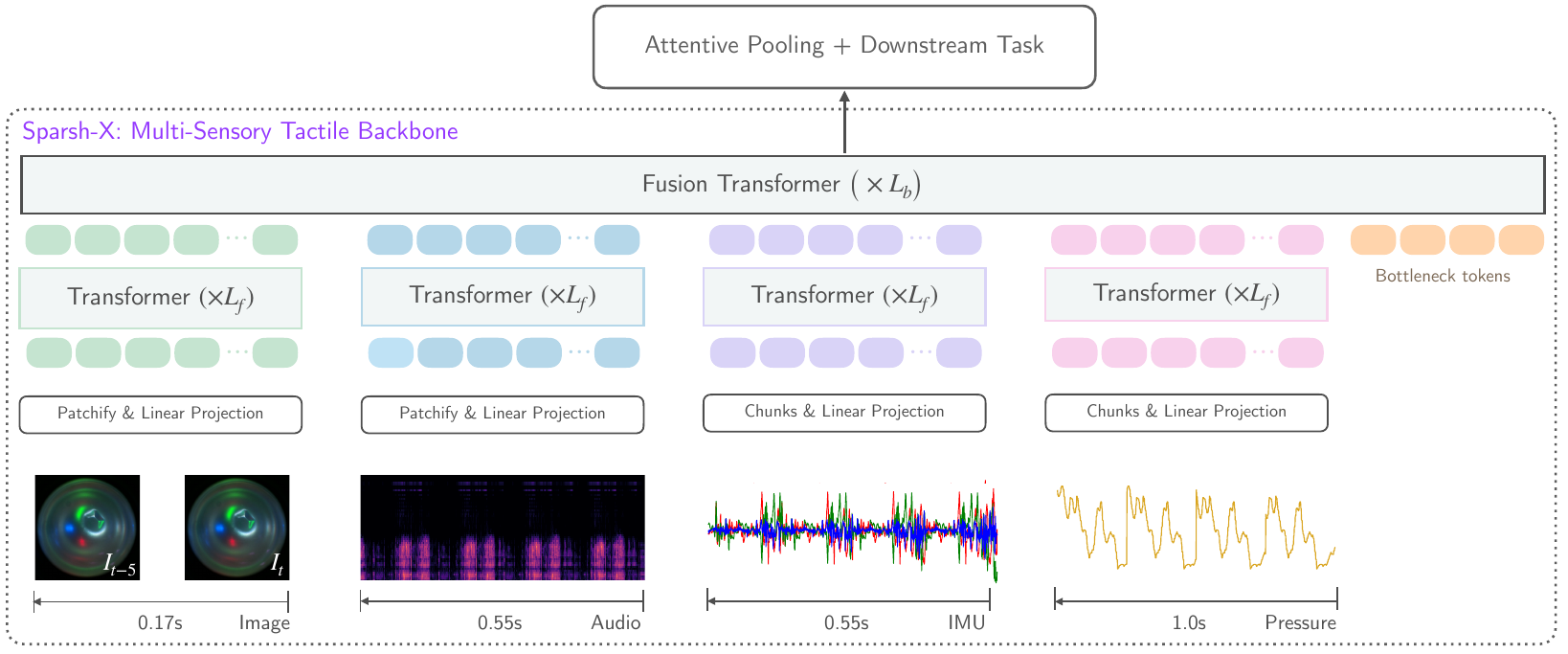}
    \vspace{-7mm}
    \caption{\model, a multisensory touch transformer for general-purpose representations, integrates four tactile inputs: image, audio, accelerometer, and pressure. Each modality is processed independently in the first $L_f$ layers, then fused using bottleneck tokens for cross-modal attention in the final $L_b$ layers.}
    \vspace{-6mm}
    \label{fig:sparsh_diagram}
\end{figure}

\textbf{Inputs and Model Architecture.} \model is a transformer-based backbone~\cite{vit} (see Figure~\ref{fig:sparsh_diagram}) where each input signal is first processed independently for $L_f$ layers through self-attention. Thereafter, we allow cross-modal information flow via attention bottlenecks, as in~\cite{NEURIPS2021_76ba9f56}. Specifically, we concatenate $B$ bottleneck fusion tokens to each modality’s embedding for the subsequent $L_b$ blocks. After each cross-modal update, the fusion tokens are averaged across modalities to promote information sharing. Intuitively, the bottleneck tokens act as multimodal summarizers, distilling and exchanging information between tactile modalities within each transformer block. Following experimental insights from~\cite{NEURIPS2021_76ba9f56}, we set the total number of transformer layers to $L = L_f + L_b = 12$, with $L_f=8$ layers for unimodal processing and $L_b=4$ fusion layers with $B=4$ bottleneck tokens.

The inputs to \model are image, audio, accelerometer, and pressure recorded by the Digit~360 sensor. 
Since all modalities have different sampling frequencies and data structures, we describe the steps for preprocessing and tokenization. \textit{Tactile images} are sampled at 30fps~\cite{higuera2024sparsh} and passed to the model with a temporal stride of 5 concatenated along the channel dimension. We crop to zoom-in the fish-eye image and resize to $224 \times 224 \times 3$. Image patches ($16 \times 16$) are then tokenized to embeddings of 768 dimensions through a linear projection layer. \textit{Audio} comes from two contact microphones sampled at 48kHz. A 0.55s window of audio signal is converted into a log-mel spectogram of 128 channels computed from a 5ms Hamming window with hop length 2.5ms. We concatenate the spectograms from both microphones, resulting into an audio input of $224 \times 256$ which is further tokenized with a patch size of 16. \textit{IMU} data from the 3-axis accelerometer is sampled at 400Hz  and combined in a window of  0.55s. The \textit{pressure} signal is sampled at 200Hz and combined in a window of 1.1s window. Both signals are  tokenized resulting in $224 \times 3$ and $224 \times 1$ temporal signals. 

 \begin{wrapfigure}{r}{5cm}
    \centering 
    \vspace{-6mm}
   \includegraphics[width=\linewidth]{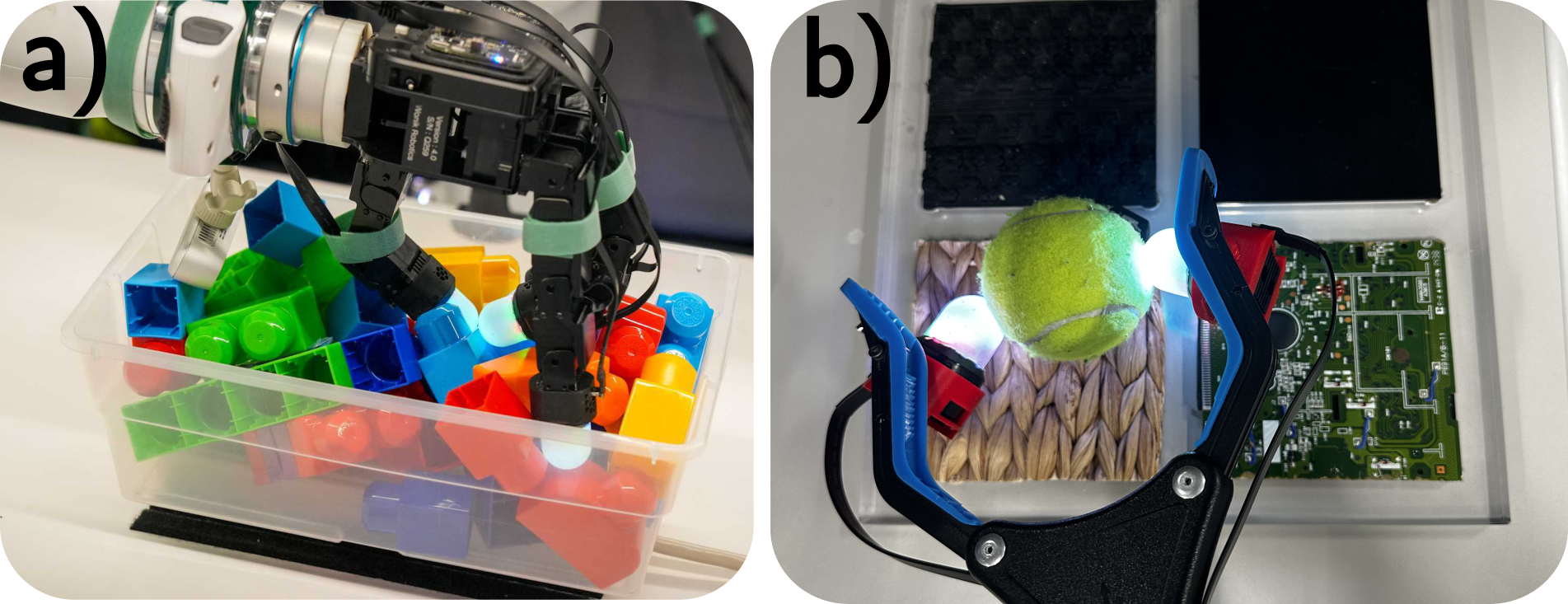}
    \caption{Pretraining data collection setup using (a) Digit~360 and Allegro hand (b) two-fingered manual picker.}
    \label{fig:pretraining-sources}
    \vspace{-5mm}
 \end{wrapfigure}

\vspace{-3mm}
\paragraph{SSL Training Pipeline.} We train \model using Self-Supervised Learning (SSL) which offers several benefits, including the ability to learn general representations, robustness to distractors, and independence from labeled data. Our SSL training dataset consists of $\ssim1$M samples  generated from two primary sources: an Allegro hand with Digit~360 sensors on the fingertips that performs random motions with objects such as dipping into a tray filled with various items; and a manual picker~\cite{chi2024universal, pmlr-v155-young21a, shafiullah2023bringing} with the same sensor adapted to the gripping mechanism, used to execute atomic manipulation actions such as picking up, sliding, tapping, placing, and dropping objects against diverse surfaces that vary in roughness, hardness, softness, friction, and texture properties. 
We employ a teacher-student self-distillation approach~\cite{caron2021emerging, oquab2023dinov2}, where both branches consist of an encoder and a predictor head. After tokenizing each multisensory touch input, appending a register token, and adding sinusoidal positional embeddings, we apply masking to the student input tokens per modality, retaining 10-50\% of the signal for local masks and 50-100\% for global masks. We concatenate the register tokens from global and local masks and pass them through their corresponding prediction heads. As in~\cite{caron2021emerging}, the prediction task involves clustering, where teacher tokens serve as pseudo-labels for the student network, with centroids that adapt over time as the model learns. We use cross-entropy between the softmax outputs of the teacher and student networks as optimization objective. We train \model for 200 epochs on 16 A-100 GPUs, with 128 batch size and AdamW optimizer with  linear rampup followed by a cosine schedule as the learning scheduler. Please refer to the Appendix for further pretraining details.


\vspace{-3mm}\section{Integrating Multisensory Touch Representations in Downstream Tasks}\vspace{-2mm}
\label{sec:sparsh_tasks}

We propose a set of downstream tasks for evaluating the capability and generalization of our representations on touch-centric tasks. Our study is driven by two central research questions, first, \textit{what physical properties do our representations capture from contact interactions?}, and second \textit{how can real-world touch representations be leveraged for manipulation policy learning?}. 
\vspace{-2mm}

\subsection{Inferring physical properties with \model}
\label{sec:sparsh_d360_sl_tasks}
\vspace{-2mm}
\begin{figure}
    \centering
    \includegraphics[width=\linewidth]{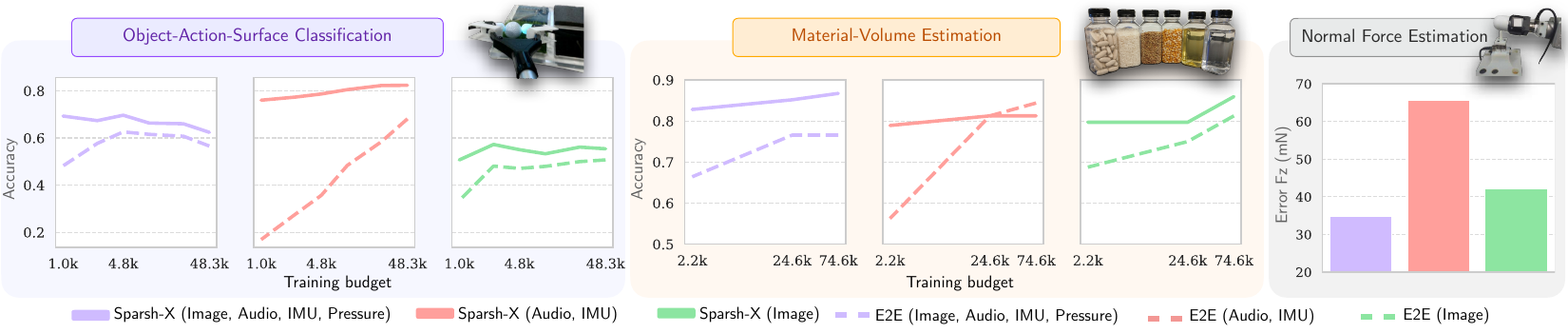}
    \caption{Performance of frozen \model representations with different tactile inputs. The synergy of multiple modalities improves object-action-surface identification (\textit{left}) and material-quantity estimation (\textit{middle}), outperforming tactile image alone and showing data efficiency over E2E. Combined modalities also enhance normal force estimation (\textit{right}), a task typically addressed with vision-based tactile sensing.}
    \vspace{-5mm}
    \label{fig:results_interpretability}
\end{figure}

We design supervised tasks to evaluate \model representations' ability to capture physical properties for robotic manipulation. These tasks cover (1) object-level characteristics (e.g., type and quantity), (2) static contact properties (e.g., force and material), and (3) dynamic interaction cues (e.g., sliding, tapping). For each task, we train a task-specific attentive decoder~\cite{chen2024context} in a supervised manner, using as input \model representations. Importantly, the encoder weights of \model are frozen to isolate and assess the quality of the representations learned from self-supervised pretraining.


\textbf{Object-Action-Surface Classification.} We train the decoder on the subset of the pre-training dataset that uses the manual picker to jointly classify the object being grasped (golf ball, LEGO, or wood block), the action being performed (planar sliding, circular sliding, or tapping), and the extrinsic surface in contact with the object (plastic, fabric, grass, or formwork). This task probes whether the representations encode static contact properties like friction, stiffness, and roughness, as well as dynamic cues that distinguish between different motion patterns. To assess the benefits of multisensory touch, we ablate the input modalities to \model and compare performance against traditional end-to-end (encoder-decoder) models trained from scratch.  As shown in Figure~\ref{fig:results_interpretability}~(left), combining tactile modalities significantly boosts accuracy. For instance, pairing audio with IMU yields a 32\% improvement, while using all modalities together provides a 13\% gain compared to using tactile images alone. Pre-training further enhances performance when using all modalities, consistently outperforming E2E models with task-specific embeddings, with a particularly notable 10\% margin under the lowest data regime.

\textbf{Material-Quantity Estimation.} We evaluate the capacity of our model to distinguish materials (both solids and liquids) and to provide a coarse mass estimation through shaking motions~\cite{Huang-RSS-22, 9197063, 8968303, pmlr-v87-clarke18a}. We train an attentive decoder to jointly classify material type and quantity by shaking 8oz bottles with a parallel gripper equipped with Digit 360 sensors (see Figure~\ref{fig:teaser}). The dataset includes four solids (corn kernels, lentils, vitamin pills, rice), two liquids with distinct viscosity (water, oil), and three fill levels (full, half, quarter). As shown in Figure~\ref{fig:results_interpretability}~(middle), \model representations from all modalities achieve the highest accuracy across all training data budgets, outperforming end-to-end models trained on tactile images alone by 20.5\%. \model representations outperform end-to-end models across all sensory inputs, demonstrating superior data efficiency and generalization.

\textbf{Normal Force Estimation.} Following the protocol from~\cite{higuera2024sparsh}, we use a hemispherical probe to apply normal forces of up to 3.5N to the Digit 360 sensor by indenting perpendicularly into the elastomer surface. An attentive regression head is trained to estimate the applied normal force from frozen \model representations across various tactile sensory inputs. Evaluation is performed on samples with randomized force magnitudes and indentation locations, using the same probe geometry. As shown in Figure~\ref{fig:results_interpretability}~(right), combining all tactile modalities leads to improvement in force estimation accuracy, achieving an average error of 35mN, a 17\% improvement over using only tactile image.




\vspace{-2mm}
\subsection{\model for Policy Learning}
\vspace{-2mm}
\label{sec:sparsh_d360_policy}

We investigate how manipulation policies can benefit from touch representations that capture physical properties like friction, mass, and forces. We demonstrate their effectiveness in real-world  (a) imitation learning and (b) tactile adaptation of sim-trained policies, evaluating on two contact-rich tasks: plug insertion and in-hand rotation.

\subsubsection*{Plug-Insertion via Imitation Learning}\vspace{-1mm}
Insertion is a fundamental skill in robot manipulation and has long served as a benchmark task in the literature \cite{dong2021tactile, sharma2025sparshskin, tang2023industreal, wu2024tacdiffusion}. We evaluate the utility of multisensory touch representations in a plug-insertion task, where a robot equipped with an Allegro hand and Digit 360 sensors must insert a pre-grasped plug into a fixed socket. Using kinesthetic teleoperation, we collect 100 demonstrations with randomized initial arm poses, recording joint states, wrist poses, camera images, and tactile data.

Our policy architecture is adapted from ACT~\cite{zhao2023learning}. The inputs include wrist camera embeddings, obtained by training a vision encoder, and \model representations for the thumb, index, and middle fingers. These tactile features are aggregated using an attentive pooling layer~\cite{chen2024context}. The model predicts a trajectory of absolute end-effector poses over a horizon of $H=8$.

\begin{wrapfigure}{r}{5.5cm}
    \centering
    \vspace{-5mm}
    \includegraphics[width=\linewidth]{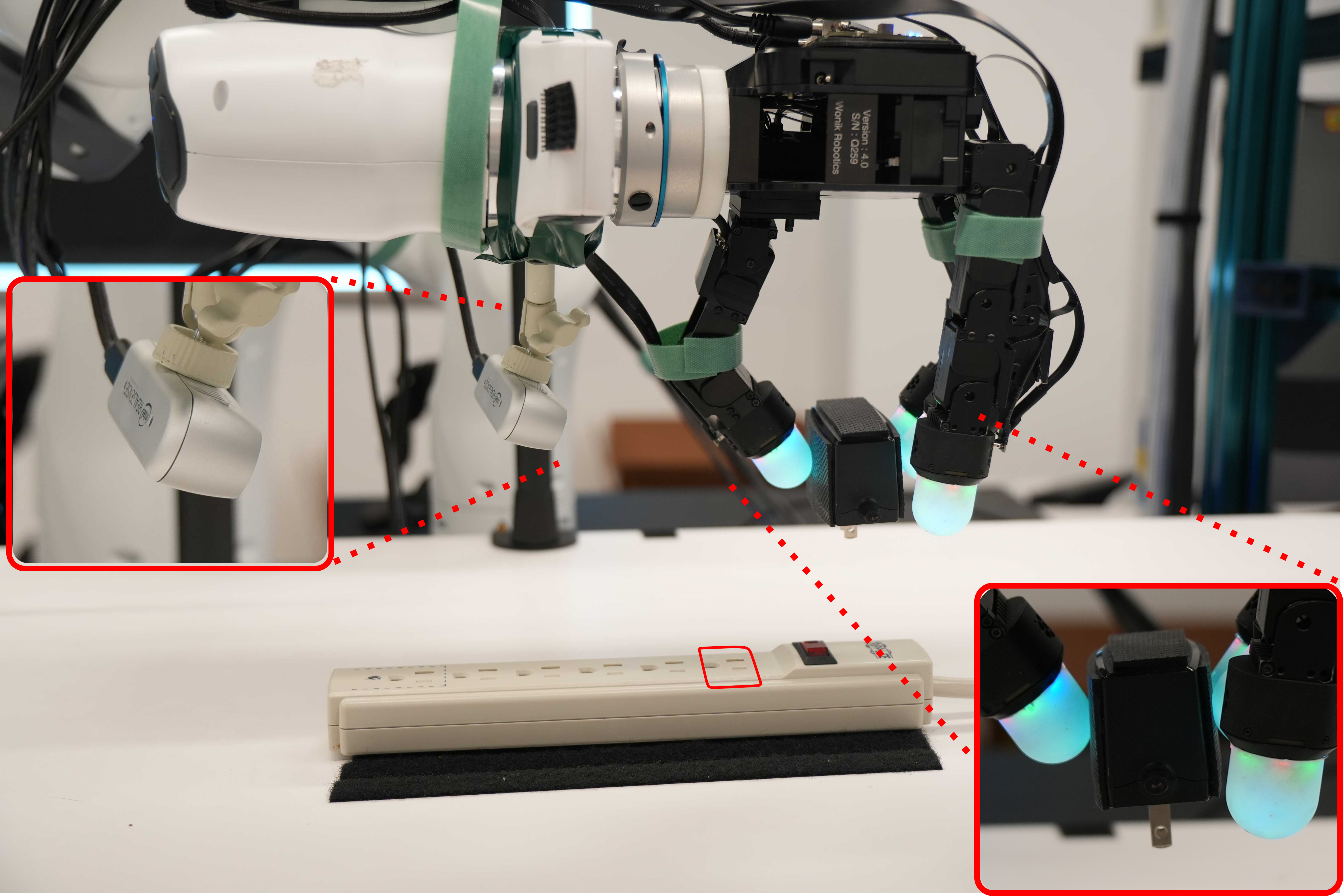}
    \includegraphics[width=\linewidth]{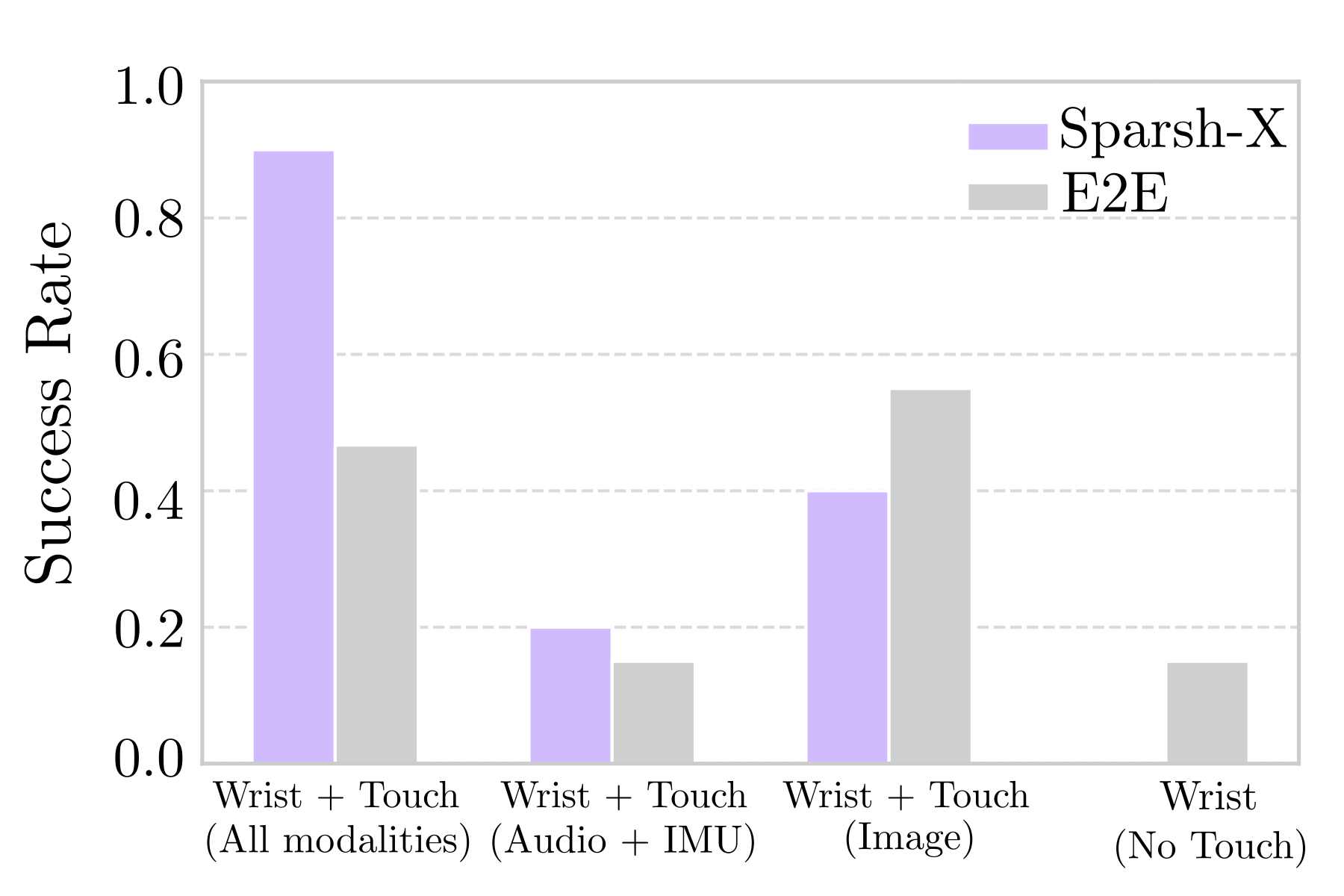}
     \caption{\textbf{Top.} Experimental setup for plug-insertion. \textbf{Bottom.} Success rate over 20 trials using different tactile sensory modes. Leveraging multimodal touch with \model improves performance by 500\% over external-vision-only and 63\% over E2E tactile-vision-only policies.}
     \vspace{-7mm}
    \label{fig:plug-insertion}
\end{wrapfigure}



\textbf{Evaluation.} To evaluate the contribution of multisensory touch, we ablate the policy by varying the combination of tactile sensory inputs used, always in conjunction with the wrist camera. We also include a vision-only baseline without touch. Each policy is evaluated over 20 trials, with randomized wrist starting positions. 


As shown in Figure~\ref{fig:plug-insertion}, multisensory touch is key for achieving high performance in a tight-tolerance insertion task, with a 90\% success rate. Pretraining plays a crucial role, yielding a ~90\% performance boost compared to training representations with all modalities from scratch jointly with the policy model on task-specific data. These results highlight the benefits of both multimodality and pretraining. Access to multiple sensing modalities enables better discrimination of subtle contact cues. For example, audio can signal initial contact or collisions, while tactile images and pressure provide information about normal and shear forces that are critical for alignment and insertion.

Policies using tactile images outperform those relying solely on audio and motion cues. Interestingly, for tactile images, end-to-end training outperforms using frozen pretrained representations. The tactile image signal varies little between trials, allowing a specialized encoder to focus on subtle changes in contact patch location and size. It is worth noting, however, that these encoders are evaluated in-distribution, suggesting that pretrained representations may still benefit from increased data diversity and scale or fine-tuning. Notably, our results show a 63\% improvement in performance when using \model with all modalities compared to an end-to-end policy trained with tactile image alone. While pretrained tactile image encoders may benefit from broader data diversity, combining touch modalities helps mitigate such limitations, given their complementary nature. As a final remark, incorporating touch significantly improves performance on this task. In particular, the wrist-only policy often fails due to visual aliasing, where insufficient camera parallax results in plug pegs incorrectly appearing directly on top of socket openings, leading to failed insertions.

\vspace{-2mm}\subsubsection*{In-Hand rotation with sim-to-real tactile adaption}

A common strategy for learning dexterous manipulation policies is to first train a base policy using privileged information typically available only in simulation (e.g., object physical properties like mass and friction, or contact signals like location and forces) and later distill the information into another model using inputs only accessible in the real-world such as proprioception~\cite{kumar2021rma}. This raises a natural question: \textit{when richer information becomes available in the real world, how can we bring policies trained in simulation closer to the privileged information setting?}

We explore sim-to-real tactile adaptation in the context of in-hand rotation. As our base policy, we use Hora~\cite{qi2022hand}, a proprioception-only policy for rotating objects along the \textit{z}-axis. Hora is trained in simulation via rapid motor adaptation~\cite{kumar2021rma}, leveraging privileged information such as object pose, shape, mass, friction, and other properties that can be perceived through touch at the fingertips.  Since \model captures physical properties, our goal is to do \emph{tactile adaptation} on top of Hora, improving stability during rotation by reducing slip.

\begin{figure}
    \centering
    \includegraphics[width=\linewidth]{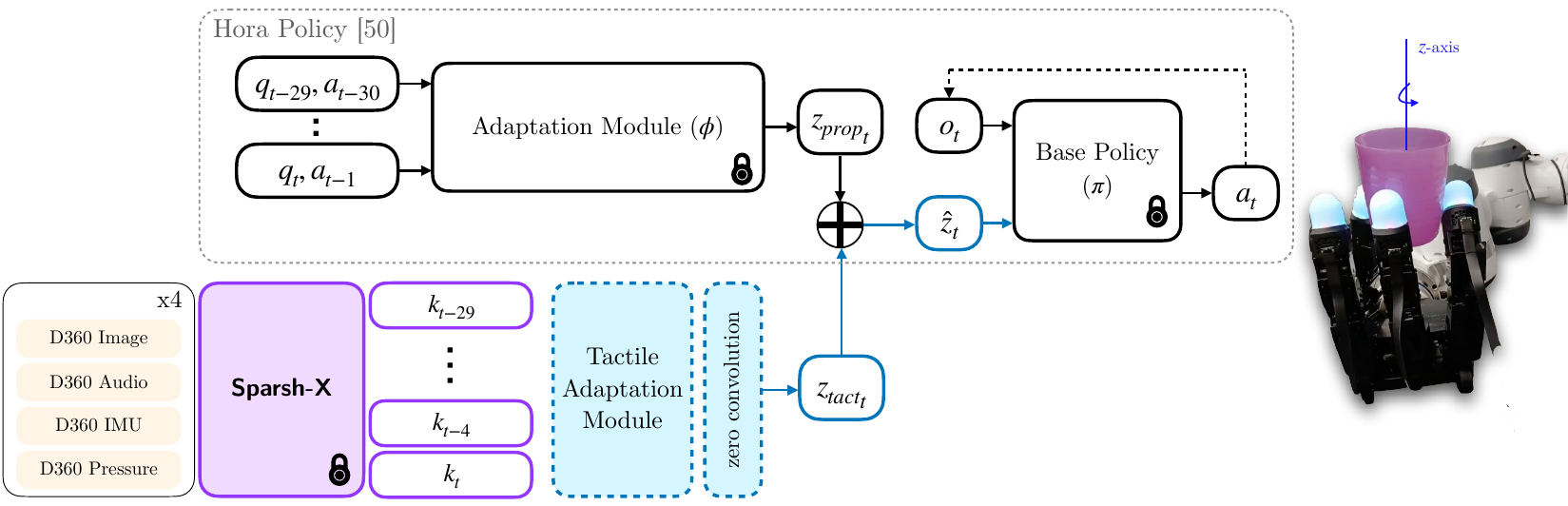}
    \caption{We introduce real-world tactile adaptation of sim-trained policies via ControlNet~\cite{Zhang_2023_ICCV}, where the zero-convolution layer enables gradual fine-tuning of the embedding $\hat{z}_t$ using \model representations.}
    \label{fig:controlnet_diagram}
    \vspace{-10mm}
\end{figure}

We propose \emph{tactile adaptation} via ControlNet~\cite{Zhang_2023_ICCV}, which allows the integration of new control modalities without retraining the base model. Applied to policy learning, ControlNet ensures that performance does not degrade below that of the original Hora policy. As shown in Figure \ref{fig:controlnet_diagram}, we learn a tactile adaptation module that connects to the frozen base policy through a zero-initialized convolutional layer, enabling progressive integration of tactile information. Specifically, we feed to the tactile adaptation module \model representations of all four fingertips over the past 1.5 seconds, aligning with the proprioceptive history window used by the base policy.



\textbf{Training.} We rollout the baseline Hora policy (using the open-sourced policy checkpoint) to collect real-world sequences, capturing proprioceptive joint states, target actions, and tactile data related to the in-hand rotation of cup-like objects. For training, we select 50 successful trajectories in which the object remains stably rotating along the \textit{z}-axis for at least 30 seconds. The tactile adaptation module is then optimized to minimize the L2 loss between the real-world joint angles and the target action output from Hora with the ControlNet.


\textbf{Evaluation.} We compare the baseline Hora policy against two tactile adaptation variants: Hora+ControlNet(\model), using pre-trained \model representations with different combinations of tactile modalities, and Hora+ControlNet(E2E), trained end-to-end. To isolate the impact of tactile feedback over improvements from rejection sampling, we also evaluate against Hora fine-tuned on real-world data and a proprioception-only imitation learning baseline. The primary goal is to improve stability during in-hand rotation by reducing slip, where the object either shifts into the palm or is completely dropped. We evaluate each policy based on vertical drift and time-to-fall, performing 10 trials per policy with a maximum episode duration of 60 seconds.

\begin{figure}
    \centering
    \includegraphics[width=\linewidth]{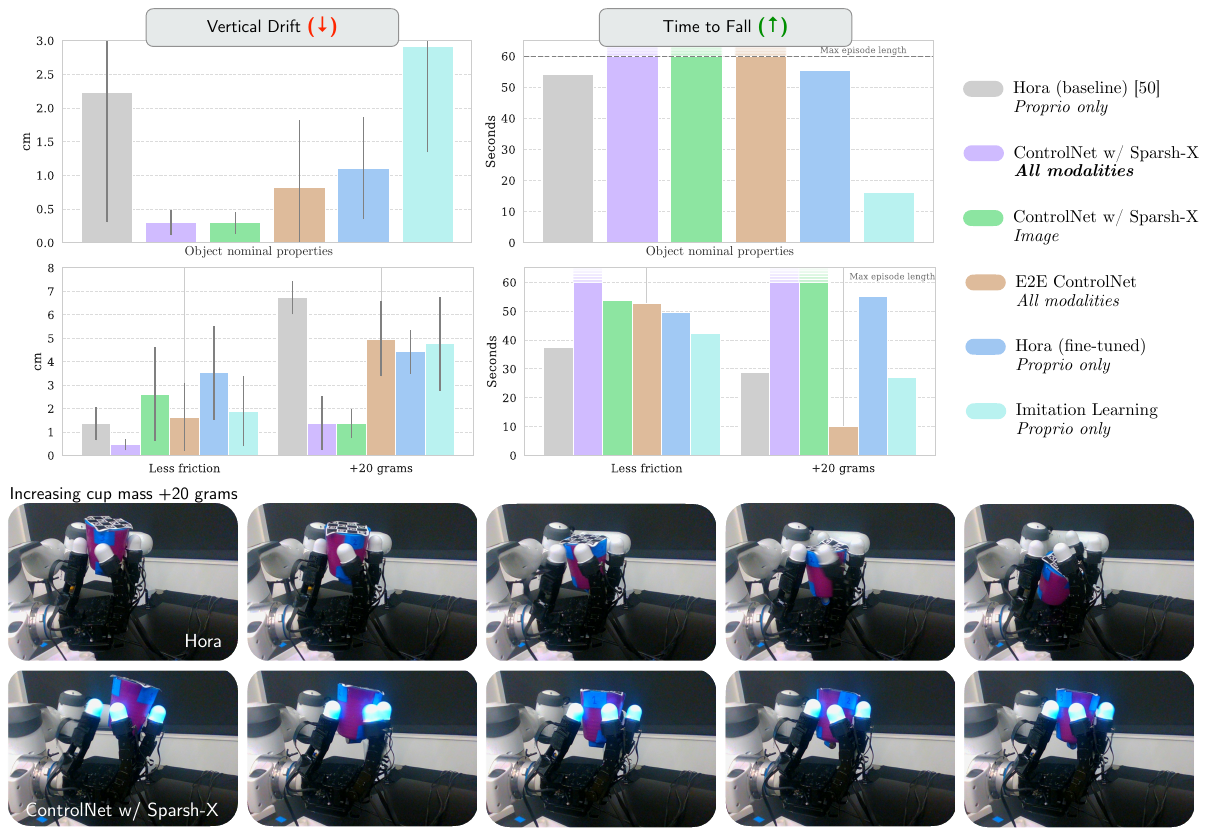}
    \caption{\textbf{Top.} For object nominal properties, tactile adaptation with \model reduces vertical drift by 90\% compared to Hora. Fine-tuning with successful rollouts does not yield same performance, highlighting the effectiveness of tactile adaptation. Metrics from 10 trials (60s episodes). \textbf{Middle.} Under dynamical changes, tactile adaptation shows superior stability than Hora variants. Metrics from 5 trials (60s episodes). \textbf{Bottom.} Snapshots of policy rollouts with and without multisensory touch input as object mass increases.}
    \label{fig:results_controlnet_results}
    \vspace{-6mm}
\end{figure}

Our ControlNet approach with \model representations, reduced vertical translation by 90\% (see Figure~\ref{fig:results_controlnet_results}~(top)), 
using either all modalities or just tactile images. 
While Hora+ControlNet(E2E) also improved stability a bit, it slowed rotation. Critically, our method outperformed both finetuned Hora and the proprioception-only imitation learning baseline, demonstrating that the benefit stems from tactile feedback, not just good demonstrations. The imitation learning baseline was unreliable, frequently failing due to out-of-distribution states as reflected in the lowest time-to-fall metric. 



We also evaluate model robustness to altering the physical properties of the object, specifically friction and mass as shown in Figure~\ref{fig:results_controlnet_results}~(middle). When friction is reduced, Hora+ControlNet(\model) outperforms all other policy variants. By using a synergy of all tactile modes, it can maintain object stability without losing grasp. In contrast, Hora+ControlNet(image) struggles, possibly indicating that changes in the contact patch are too subtle to be captured by \model from tactile images alone. This result underscores the complementary strengths of multisensory touch.  When the object's mass is increased, Hora+ControlNet(\model) and Hora+ControlNet(image) allow the baseline policy to adapt its finger gaiting to compensate for the added weight. This adaptation is possible because the tactile properties of the object can be captured by \model and transferred to the privilege information embedding in the latent space.

\vspace{-3mm}\section{Discussion and Conclusion}
\label{sec:conclusion}

\vspace{-4mm}

We present \model, a self-supervised backbone for general multisensory touch representations. Through both policy learning and supervised tactile experiments, we demonstrate that incorporating multisensory touch from Digit 360 and scale pretraining over $\sim1M$ samples significantly enhances task performance compared to using end-to-end approaches with tactile images alone.

Our study is driven by two central research questions. First, \textit{how can real-world touch representations be leveraged for manipulation policy learning?} \model pretraining enables better and more robust policies. We explore two approaches: imitation learning (IL) and tactile adaptation of sim-trained policies. IL is naturally suited as demonstrations capture rich tactile signals from all modalities provided by Digit 360. For sim-policies, we propose tactile adaptation via ControlNet, enabling the propagation of tactile information previously accessible only in simulation as privileged information. We validate these approaches on two fundamental manipulation tasks: plug insertion and in-hand object rotation. Notably, \model enhances policy performance by 63\% over policies using tactile images alone and improves robustness by 90\% by using touch to recover object state during manipulation.

Second, \textit{what tactile properties do our representations capture?} We find that \model representations effectively captures physical properties that allows to identify objects, actions, surfaces, estimate intrinsics properties and forces from multisensory touch signals. We evaluate \model on a suite of supervised benchmark tasks common in the literature: object-action recognition, material-quantity estimation, and force prediction. We perform ablations over tactile input modalities and training data budgets to assess the impact of pretraining and multisensory fusion. Our results show that a synergy of touch from images, audio, IMU, and pressure,  leads to higher accuracy across all tasks even in low-data regimes. Compared to training end-to-end with tactile images only, \model achieves an average improvement of 48\% across all tasks, demonstrating the benefits of both pretraining and touch sensory fusion.

\newpage
\section*{Limitations}
\label{sec:limitations}
While our study highlights the benefits of multisensory touch and self-supervised pretraining, some limitations remain. Each modality introduces its own challenges in capturing a broad and diverse set of contact interactions at scale. In our pretraining dataset, the tactile image modality from the Digit 360 sensor exhibits the lowest diversity in terms of number of different devices used with their own optical artifacts, potentially limiting its generalization in downstream performance. We believe that as the community increasingly adopts multisensory tactile sensors like the Digit 360, collaborative efforts can help build larger and more diverse datasets to support scalable pretraining. Additionally, our experiments focus exclusively on frozen \model representations to understand the pure impact of pretraining on generalization. However, allowing fine-tuning with task-specific data could further improve performance and help compensate for modality-specific data limitations. Finally, our evaluation of force sensing is limited to normal force estimation under controlled contact conditions. Generalizing to varying contact geometries and multiple simultaneous contacts remains an open area for future work. Moreover, we do not consider shear force estimation in this study, as separating the effects of extrinsic forces from the internal deformation of the elastomer presents non-trivial modeling challenges.

\acknowledgments{The authors thank Youngsun Wi, Changhao Wang, Haozhi Qi, Luis Pineda, Jessica Yin, Tarasha Khurana and Jitendra Malik for helpful discussions on the research and reviews of the paper. This work is supported by Meta FAIR labs.}


\bibliography{references}  

\begin{thebibliography}{54}
\providecommand{\natexlab}[1]{#1}
\providecommand{\url}[1]{\texttt{#1}}
\expandafter\ifx\csname urlstyle\endcsname\relax
  \providecommand{\doi}[1]{doi: #1}\else
  \providecommand{\doi}{doi: \begingroup \urlstyle{rm}\Url}\fi

\bibitem[Yuan et~al.(2017)Yuan, Dong, and Adelson]{gelsight}
W.~Yuan, S.~Dong, and E.~H. Adelson.
\newblock Gel{S}ight: High-resolution robot tactile sensors for estimating geometry and force.
\newblock \emph{Sensors}, 17\penalty0 (12), 2017.
\newblock ISSN 1424-8220.
\newblock \doi{10.3390/s17122762}.
\newblock URL \url{https://www.mdpi.com/1424-8220/17/12/2762}.

\bibitem[Lambeta et~al.(2020)Lambeta, Chou, Tian, Yang, Maloon, Most, Stroud, Santos, Byagowi, Kammerer, Jayaraman, and Calandra]{digit}
M.~Lambeta, P.-W. Chou, S.~Tian, B.~Yang, B.~Maloon, V.~R. Most, D.~Stroud, R.~Santos, A.~Byagowi, G.~Kammerer, D.~Jayaraman, and R.~Calandra.
\newblock Digit: A novel design for a low-cost compact high-resolution tactile sensor with application to in-hand manipulation.
\newblock \emph{IEEE Robotics and Automation Letters}, 5\penalty0 (3):\penalty0 3838--3845, 2020.
\newblock \doi{10.1109/LRA.2020.2977257}.

\bibitem[Lepora(2021)]{lepora2021soft}
N.~F. Lepora.
\newblock Soft biomimetic optical tactile sensing with the tactip: A review.
\newblock \emph{IEEE Sensors Journal}, 21\penalty0 (19):\penalty0 21131--21143, 2021.

\bibitem[Lambeta et~al.(2024)Lambeta, Wu, Sengul, Most, Black, Sawyer, Mercado, Qi, Sohn, Taylor, Tydingco, Kammerer, Stroud, Khatha, Jenkins, Most, Stein, Chavira, Craven-Bartle, Sanchez, Ding, Malik, and Calandra]{lambeta2024d360}
M.~Lambeta, T.~Wu, A.~Sengul, V.~R. Most, N.~Black, K.~Sawyer, R.~Mercado, H.~Qi, A.~Sohn, B.~Taylor, N.~Tydingco, G.~Kammerer, D.~Stroud, J.~Khatha, K.~Jenkins, K.~Most, N.~Stein, R.~Chavira, T.~Craven-Bartle, E.~Sanchez, Y.~Ding, J.~Malik, and R.~Calandra.
\newblock Digitizing touch with an artificial multimodal fingertip, 2024.
\newblock URL \url{https://arxiv.org/abs/2411.02479}.

\bibitem[Yu et~al.(2024)Yu, Han, Wang, Saxena, Xu, and Zhao]{yu2024mimictouch}
K.~Yu, Y.~Han, Q.~Wang, V.~Saxena, D.~Xu, and Y.~Zhao.
\newblock Mimictouch: Leveraging multi-modal human tactile demonstrations for contact-rich manipulation.
\newblock In \emph{8th Annual Conference on Robot Learning}, 2024.
\newblock URL \url{https://openreview.net/forum?id=7yMZAUkXa4}.

\bibitem[Mejia et~al.(2024)Mejia, Dean, Hellebrekers, and Gupta]{mejia2024hearing}
J.~Mejia, V.~Dean, T.~Hellebrekers, and A.~Gupta.
\newblock Hearing touch: Audio-visual pretraining for contact-rich manipulation.
\newblock In \emph{2024 IEEE International Conference on Robotics and Automation (ICRA)}, pages 6912--6919. IEEE, 2024.

\bibitem[Xu et~al.(2025)Xu, Uppuluri, Zhang, Fitch, Crandall, Shou, Wang, and She]{xu2025unit}
Z.~Xu, R.~Uppuluri, X.~Zhang, C.~Fitch, P.~G. Crandall, W.~Shou, D.~Wang, and Y.~She.
\newblock {UniT}: Data efficient tactile representation with generalization to unseen objects, 2025.
\newblock URL \url{https://arxiv.org/abs/2408.06481}.

\bibitem[Zhao et~al.(2024)Zhao, Ma, Wang, and Adelson]{zhao2024transferable}
J.~Zhao, Y.~Ma, L.~Wang, and E.~Adelson.
\newblock Transferable tactile transformers for representation learning across diverse sensors and tasks.
\newblock In \emph{8th Annual Conference on Robot Learning}, 2024.
\newblock URL \url{https://openreview.net/forum?id=KXsropnmNI}.

\bibitem[Higuera et~al.(2024)Higuera, Sharma, Bodduluri, Fan, Lancaster, Kalakrishnan, Kaess, Boots, Lambeta, Wu, and Mukadam]{higuera2024sparsh}
C.~Higuera, A.~Sharma, C.~K. Bodduluri, T.~Fan, P.~Lancaster, M.~Kalakrishnan, M.~Kaess, B.~Boots, M.~Lambeta, T.~Wu, and M.~Mukadam.
\newblock Sparsh: Self-supervised touch representations for vision-based tactile sensing.
\newblock In \emph{8th Annual Conference on Robot Learning}, 2024.
\newblock URL \url{https://openreview.net/forum?id=xYJn2e1uu8}.

\bibitem[Gupta et~al.(2025)Gupta, Mo, Jin, and Yuan]{gupta2025sensor}
H.~Gupta, Y.~Mo, S.~Jin, and W.~Yuan.
\newblock Sensor-invariant tactile representation.
\newblock \emph{arXiv preprint arXiv:2502.19638}, 2025.

\bibitem[Oquab et~al.(2023)Oquab, Darcet, Moutakanni, Vo, Szafraniec, Khalidov, Fernandez, Haziza, Massa, El-Nouby, et~al.]{oquab2023dinov2}
M.~Oquab, T.~Darcet, T.~Moutakanni, H.~Vo, M.~Szafraniec, V.~Khalidov, P.~Fernandez, D.~Haziza, F.~Massa, A.~El-Nouby, et~al.
\newblock {DINOv2}: Learning robust visual features without supervision.
\newblock \emph{arXiv preprint arXiv:2304.07193}, 2023.

\bibitem[Bardes et~al.(2023)Bardes, Garrido, Ponce, Chen, Rabbat, LeCun, Assran, and Ballas]{bardes2023vjepa}
A.~Bardes, Q.~Garrido, J.~Ponce, X.~Chen, M.~Rabbat, Y.~LeCun, M.~Assran, and N.~Ballas.
\newblock V-{JEPA}: Latent video prediction for visual representation learning.
\newblock 2023.

\bibitem[Hendrycks et~al.(2019)Hendrycks, Mazeika, Kadavath, and Song]{hendrycks2019using}
D.~Hendrycks, M.~Mazeika, S.~Kadavath, and D.~Song.
\newblock Using self-supervised learning can improve model robustness and uncertainty.
\newblock \emph{Advances in neural information processing systems}, 32, 2019.

\bibitem[Karnan et~al.(2023)Karnan, Yang, Farkash, Warnell, Biswas, and Stone]{karnan2023sterling}
H.~Karnan, E.~Yang, D.~Farkash, G.~Warnell, J.~Biswas, and P.~Stone.
\newblock Sterling: Self-supervised terrain representation learning from unconstrained robot experience.
\newblock \emph{arXiv preprint arXiv:2309.15302}, 2023.

\bibitem[Lin et~al.(2022)Lin, Lloyd, Church, and Lepora]{lin2022tactile}
Y.~Lin, J.~Lloyd, A.~Church, and N.~F. Lepora.
\newblock Tactile gym 2.0: Sim-to-real deep reinforcement learning for comparing low-cost high-resolution robot touch.
\newblock \emph{IEEE Robotics and Automation Letters}, 7\penalty0 (4):\penalty0 10754--10761, 2022.

\bibitem[Donlon et~al.(2018)Donlon, Dong, Liu, Li, Adelson, and Rodriguez]{8593661}
E.~Donlon, S.~Dong, M.~Liu, J.~Li, E.~Adelson, and A.~Rodriguez.
\newblock Gelslim: A high-resolution, compact, robust, and calibrated tactile-sensing finger.
\newblock In \emph{2018 IEEE/RSJ International Conference on Intelligent Robots and Systems (IROS)}, pages 1927--1934, 2018.
\newblock \doi{10.1109/IROS.2018.8593661}.

\bibitem[Yuan et~al.(2018)Yuan, Mo, Wang, and Adelson]{8461164}
W.~Yuan, Y.~Mo, S.~Wang, and E.~H. Adelson.
\newblock Active clothing material perception using tactile sensing and deep learning.
\newblock In \emph{2018 IEEE International Conference on Robotics and Automation (ICRA)}, pages 4842--4849, 2018.
\newblock \doi{10.1109/ICRA.2018.8461164}.

\bibitem[Guo et~al.(2023)Guo, Huang, and Yuan]{10341880}
X.~Guo, H.-J. Huang, and W.~Yuan.
\newblock Estimating properties of solid particles inside container using touch sensing.
\newblock In \emph{2023 IEEE/RSJ International Conference on Intelligent Robots and Systems (IROS)}, pages 8985--8992, 2023.
\newblock \doi{10.1109/IROS55552.2023.10341880}.

\bibitem[Suresh et~al.(2024)Suresh, Qi, Wu, Fan, Pineda, Lambeta, Malik, Kalakrishnan, Calandra, Kaess, et~al.]{suresh2024neuralfeels}
S.~Suresh, H.~Qi, T.~Wu, T.~Fan, L.~Pineda, M.~Lambeta, J.~Malik, M.~Kalakrishnan, R.~Calandra, M.~Kaess, et~al.
\newblock Neuralfeels with neural fields: Visuotactile perception for in-hand manipulation.
\newblock \emph{Science Robotics}, 9\penalty0 (96):\penalty0 eadl0628, 2024.

\bibitem[Gao et~al.(2024)Gao, Deng, Yang, Yuan, and Zhu]{gao2024exploiting}
R.~Gao, K.~Deng, G.~Yang, W.~Yuan, and J.-Y. Zhu.
\newblock Tactile dreamfusion: Exploiting tactile sensing for 3d generation.
\newblock In \emph{Conference on Neural Information Processing Systems (NeurIPS)}, 2024.

\bibitem[Suresh et~al.(2023)Suresh, Si, Anderson, Kaess, and Mukadam]{suresh2023midastouch}
S.~Suresh, Z.~Si, S.~Anderson, M.~Kaess, and M.~Mukadam.
\newblock Midastouch: Monte-carlo inference over distributions across sliding touch.
\newblock In \emph{Conference on Robot Learning}, pages 319--331. PMLR, 2023.

\bibitem[Huang et~al.(2024)Huang, Kaess, and Yuan]{huang2024normalflow}
H.-J. Huang, M.~Kaess, and W.~Yuan.
\newblock Normalflow: Fast, robust, and accurate contact-based object 6dof pose tracking with vision-based tactile sensors.
\newblock \emph{IEEE Robotics and Automation Letters}, 2024.

\bibitem[Dong et~al.(2021)Dong, Jha, Romeres, Kim, Nikovski, and Rodriguez]{dong2021tactile}
S.~Dong, D.~K. Jha, D.~Romeres, S.~Kim, D.~Nikovski, and A.~Rodriguez.
\newblock Tactile-rl for insertion: Generalization to objects of unknown geometry.
\newblock In \emph{2021 IEEE International Conference on Robotics and Automation (ICRA)}, pages 6437--6443. IEEE, 2021.

\bibitem[Aquilina et~al.(2024)Aquilina, Barton, and Lepora]{aquilina2024tactile}
K.~Aquilina, D.~A. Barton, and N.~F. Lepora.
\newblock Tactile control for object tracking and dynamic contour following.
\newblock \emph{Robotics and Autonomous Systems}, 178:\penalty0 104710, 2024.

\bibitem[Xue et~al.(2025)Xue, Ren, Chen, Zhang, Fang, Gu, Xu, and Lu]{xue2025reactive}
H.~Xue, J.~Ren, W.~Chen, G.~Zhang, Y.~Fang, G.~Gu, H.~Xu, and C.~Lu.
\newblock Reactive diffusion policy: Slow-fast visual-tactile policy learning for contact-rich manipulation.
\newblock \emph{arXiv preprint arXiv:2503.02881}, 2025.

\bibitem[Gandhi et~al.(2020)Gandhi, Gupta, and Pinto]{gandhi2020swoosh}
D.~Gandhi, A.~Gupta, and L.~Pinto.
\newblock Swoosh! rattle! thump!--actions that sound.
\newblock \emph{arXiv preprint arXiv:2007.01851}, 2020.

\bibitem[Clarke et~al.(2022)Clarke, Heravi, Rau, Gao, Wu, James, and Bohg]{clarke2022diffimpact}
S.~Clarke, N.~Heravi, M.~Rau, R.~Gao, J.~Wu, D.~James, and J.~Bohg.
\newblock Diffimpact: Differentiable rendering and identification of impact sounds.
\newblock In \emph{Conference on Robot Learning}, pages 662--673. PMLR, 2022.

\bibitem[Thankaraj and Pinto(2023)]{thankaraj2023sounds}
A.~Thankaraj and L.~Pinto.
\newblock That sounds right: Auditory self-supervision for dynamic robot manipulation.
\newblock In \emph{Conference on Robot Learning}, pages 1036--1049. PMLR, 2023.

\bibitem[Liu et~al.(2024)Liu, Chi, Cousineau, Kuppuswamy, Burchfiel, and Song]{liu2024maniwav}
Z.~Liu, C.~Chi, E.~Cousineau, N.~Kuppuswamy, B.~Burchfiel, and S.~Song.
\newblock Maniwav: Learning robot manipulation from in-the-wild audio-visual data.
\newblock In \emph{8th Annual Conference on Robot Learning}, 2024.

\bibitem[Feng et~al.(2025)Feng, Hu, Xia, Gao, Shen, Sun, Fang, and Hu]{feng2025anytouch}
R.~Feng, J.~Hu, W.~Xia, T.~Gao, A.~Shen, Y.~Sun, B.~Fang, and D.~Hu.
\newblock Anytouch: Learning unified static-dynamic representation across multiple visuo-tactile sensors.
\newblock \emph{arXiv preprint arXiv:2502.12191}, 2025.

\bibitem[Gong et~al.(2021)Gong, Chung, and Glass]{gong2021ast}
Y.~Gong, Y.-A. Chung, and J.~Glass.
\newblock Ast: Audio spectrogram transformer.
\newblock \emph{arXiv preprint arXiv:2104.01778}, 2021.

\bibitem[Niizumi et~al.(2023)Niizumi, Takeuchi, Ohishi, Harada, and Kashino]{9944865}
D.~Niizumi, D.~Takeuchi, Y.~Ohishi, N.~Harada, and K.~Kashino.
\newblock Byol for audio: Exploring pre-trained general-purpose audio representations.
\newblock \emph{IEEE/ACM Transactions on Audio, Speech, and Language Processing}, 31:\penalty0 137--151, 2023.
\newblock \doi{10.1109/TASLP.2022.3221007}.

\bibitem[Morgado et~al.(2021)Morgado, Vasconcelos, and Misra]{morgado2021audio}
P.~Morgado, N.~Vasconcelos, and I.~Misra.
\newblock Audio-visual instance discrimination with cross-modal agreement.
\newblock In \emph{Proceedings of the IEEE/CVF conference on computer vision and pattern recognition}, pages 12475--12486, 2021.

\bibitem[Li et~al.(2023)Li, Zhang, Zhu, Wang, Lee, Xu, Adelson, Fei-Fei, Gao, and Wu]{pmlr-v205-li23c}
H.~Li, Y.~Zhang, J.~Zhu, S.~Wang, M.~A. Lee, H.~Xu, E.~Adelson, L.~Fei-Fei, R.~Gao, and J.~Wu.
\newblock See, hear, and feel: Smart sensory fusion for robotic manipulation.
\newblock In K.~Liu, D.~Kulic, and J.~Ichnowski, editors, \emph{Proceedings of The 6th Conference on Robot Learning}, volume 205 of \emph{Proceedings of Machine Learning Research}, pages 1368--1378. PMLR, 14--18 Dec 2023.
\newblock URL \url{https://proceedings.mlr.press/v205/li23c.html}.

\bibitem[Nagrani et~al.(2021)Nagrani, Yang, Arnab, Jansen, Schmid, and Sun]{NEURIPS2021_76ba9f56}
A.~Nagrani, S.~Yang, A.~Arnab, A.~Jansen, C.~Schmid, and C.~Sun.
\newblock Attention bottlenecks for multimodal fusion.
\newblock In M.~Ranzato, A.~Beygelzimer, Y.~Dauphin, P.~Liang, and J.~W. Vaughan, editors, \emph{Advances in Neural Information Processing Systems}, volume~34, pages 14200--14213. Curran Associates, Inc., 2021.
\newblock URL \url{https://proceedings.neurips.cc/paper_files/paper/2021/file/76ba9f564ebbc35b1014ac498fafadd0-Paper.pdf}.

\bibitem[Dosovitskiy et~al.(2020)Dosovitskiy, Beyer, Kolesnikov, Weissenborn, Zhai, Unterthiner, Dehghani, Minderer, Heigold, Gelly, et~al.]{vit}
A.~Dosovitskiy, L.~Beyer, A.~Kolesnikov, D.~Weissenborn, X.~Zhai, T.~Unterthiner, M.~Dehghani, M.~Minderer, G.~Heigold, S.~Gelly, et~al.
\newblock An image is worth 16x16 words: Transformers for image recognition at scale.
\newblock \emph{arXiv preprint arXiv:2010.11929}, 2020.

\bibitem[Chi et~al.(2024)Chi, Xu, Pan, Cousineau, Burchfiel, Feng, Tedrake, and Song]{chi2024universal}
C.~Chi, Z.~Xu, C.~Pan, E.~Cousineau, B.~Burchfiel, S.~Feng, R.~Tedrake, and S.~Song.
\newblock Universal manipulation interface: In-the-wild robot teaching without in-the-wild robots.
\newblock In \emph{Proceedings of Robotics: Science and Systems (RSS)}, 2024.

\bibitem[Young et~al.(2021)Young, Gandhi, Tulsiani, Gupta, Abbeel, and Pinto]{pmlr-v155-young21a}
S.~Young, D.~Gandhi, S.~Tulsiani, A.~Gupta, P.~Abbeel, and L.~Pinto.
\newblock Visual imitation made easy.
\newblock In J.~Kober, F.~Ramos, and C.~Tomlin, editors, \emph{Proceedings of the 2020 Conference on Robot Learning}, volume 155 of \emph{Proceedings of Machine Learning Research}, pages 1992--2005. PMLR, 16--18 Nov 2021.
\newblock URL \url{https://proceedings.mlr.press/v155/young21a.html}.

\bibitem[Shafiullah et~al.(2023)Shafiullah, Rai, Etukuru, Liu, Misra, Chintala, and Pinto]{shafiullah2023bringing}
N.~M.~M. Shafiullah, A.~Rai, H.~Etukuru, Y.~Liu, I.~Misra, S.~Chintala, and L.~Pinto.
\newblock On bringing robots home.
\newblock \emph{arXiv preprint arXiv:2311.16098}, 2023.

\bibitem[Caron et~al.(2021)Caron, Touvron, Misra, J{\'e}gou, Mairal, Bojanowski, and Joulin]{caron2021emerging}
M.~Caron, H.~Touvron, I.~Misra, H.~J{\'e}gou, J.~Mairal, P.~Bojanowski, and A.~Joulin.
\newblock Emerging properties in self-supervised vision transformers.
\newblock In \emph{Proceedings of the IEEE/CVF international conference on computer vision}, pages 9650--9660, 2021.

\bibitem[Chen et~al.(2024)Chen, Ding, Wang, Xin, Mo, Wang, Han, Luo, Zeng, and Wang]{chen2024context}
X.~Chen, M.~Ding, X.~Wang, Y.~Xin, S.~Mo, Y.~Wang, S.~Han, P.~Luo, G.~Zeng, and J.~Wang.
\newblock Context autoencoder for self-supervised representation learning.
\newblock \emph{International Journal of Computer Vision}, 132\penalty0 (1):\penalty0 208--223, 2024.

\bibitem[Huang et~al.(2022)Huang, Guo, and Yuan]{Huang-RSS-22}
H.-J. Huang, X.~Guo, and W.~Yuan.
\newblock {Understanding Dynamic Tactile Sensing for Liquid Property Estimation}.
\newblock In \emph{Proceedings of Robotics: Science and Systems}, New York City, NY, USA, June 2022.
\newblock \doi{10.15607/RSS.2022.XVIII.072}.

\bibitem[Matl et~al.(2020)Matl, Narang, Bajcsy, Ramos, and Fox]{9197063}
C.~Matl, Y.~Narang, R.~Bajcsy, F.~Ramos, and D.~Fox.
\newblock Inferring the material properties of granular media for robotic tasks.
\newblock In \emph{2020 IEEE International Conference on Robotics and Automation (ICRA)}, pages 2770--2777, 2020.
\newblock \doi{10.1109/ICRA40945.2020.9197063}.

\bibitem[Liang et~al.(2019)Liang, Li, Ma, Hendrich, Gerkmann, Sun, and Zhang]{8968303}
H.~Liang, S.~Li, X.~Ma, N.~Hendrich, T.~Gerkmann, F.~Sun, and J.~Zhang.
\newblock Making sense of audio vibration for liquid height estimation in robotic pouring.
\newblock In \emph{2019 IEEE/RSJ International Conference on Intelligent Robots and Systems (IROS)}, pages 5333--5339, 2019.
\newblock \doi{10.1109/IROS40897.2019.8968303}.

\bibitem[Clarke et~al.(2018)Clarke, Rhodes, Atkeson, and Kroemer]{pmlr-v87-clarke18a}
S.~Clarke, T.~Rhodes, C.~G. Atkeson, and O.~Kroemer.
\newblock Learning audio feedback for estimating amount and flow of granular material.
\newblock In A.~Billard, A.~Dragan, J.~Peters, and J.~Morimoto, editors, \emph{Proceedings of The 2nd Conference on Robot Learning}, volume~87 of \emph{Proceedings of Machine Learning Research}, pages 529--550. PMLR, 29--31 Oct 2018.
\newblock URL \url{https://proceedings.mlr.press/v87/clarke18a.html}.

\bibitem[Sharma et~al.(2025)Sharma, Higuera, Bodduluri, Liu, Fan, Hellebrekers, Lambeta, Boots, Kaess, Wu, Hogan, and Mukadam]{sharma2025sparshskin}
A.~Sharma, C.~Higuera, C.~K. Bodduluri, Z.~Liu, T.~Fan, T.~Hellebrekers, M.~Lambeta, B.~Boots, M.~Kaess, T.~Wu, F.~R. Hogan, and M.~Mukadam.
\newblock Self-supervised perception for tactile skin covered dexterous hands, 2025.
\newblock URL \url{https://arxiv.org/abs/2505.11420}.

\bibitem[Tang et~al.(2023)Tang, Lin, Akinola, Handa, Sukhatme, Ramos, Fox, and Narang]{tang2023industreal}
B.~Tang, M.~A. Lin, I.~Akinola, A.~Handa, G.~S. Sukhatme, F.~Ramos, D.~Fox, and Y.~Narang.
\newblock Industreal: Transferring contact-rich assembly tasks from simulation to reality.
\newblock \emph{arXiv preprint arXiv:2305.17110}, 2023.

\bibitem[Wu et~al.(2024)Wu, Chen, Wu, Chen, Zhang, Bing, Swikir, Haddadin, and Knoll]{wu2024tacdiffusion}
Y.~Wu, Z.~Chen, F.~Wu, L.~Chen, L.~Zhang, Z.~Bing, A.~Swikir, S.~Haddadin, and A.~Knoll.
\newblock Tacdiffusion: Force-domain diffusion policy for precise tactile manipulation.
\newblock \emph{arXiv preprint arXiv:2409.11047}, 2024.

\bibitem[Zhao et~al.(2023)Zhao, Kumar, Levine, and Finn]{zhao2023learning}
T.~Z. Zhao, V.~Kumar, S.~Levine, and C.~Finn.
\newblock Learning fine-grained bimanual manipulation with low-cost hardware.
\newblock \emph{arXiv preprint arXiv:2304.13705}, 2023.

\bibitem[Kumar et~al.(2021)Kumar, Fu, Pathak, and Malik]{kumar2021rma}
A.~Kumar, Z.~Fu, D.~Pathak, and J.~Malik.
\newblock Rma: Rapid motor adaptation for legged robots.
\newblock \emph{arXiv preprint arXiv:2107.04034}, 2021.

\bibitem[Qi et~al.(2022)Qi, Kumar, Calandra, Ma, and Malik]{qi2022hand}
H.~Qi, A.~Kumar, R.~Calandra, Y.~Ma, and J.~Malik.
\newblock {In-Hand Object Rotation via Rapid Motor Adaptation}.
\newblock In \emph{Conference on Robot Learning (CoRL)}, 2022.

\bibitem[Zhang et~al.(2023)Zhang, Rao, and Agrawala]{Zhang_2023_ICCV}
L.~Zhang, A.~Rao, and M.~Agrawala.
\newblock Adding conditional control to text-to-image diffusion models.
\newblock In \emph{Proceedings of the IEEE/CVF International Conference on Computer Vision (ICCV)}, pages 3836--3847, October 2023.

\bibitem[He et~al.(2016)He, Zhang, Ren, and Sun]{he2016deep}
K.~He, X.~Zhang, S.~Ren, and J.~Sun.
\newblock Deep residual learning for image recognition.
\newblock In \emph{Proceedings of the IEEE conference on computer vision and pattern recognition}, pages 770--778, 2016.

\bibitem[Cadene et~al.(2024)Cadene, Alibert, Soare, Gallouedec, Zouitine, and Wolf]{cadene2024lerobot}
R.~Cadene, S.~Alibert, A.~Soare, Q.~Gallouedec, A.~Zouitine, and T.~Wolf.
\newblock Lerobot: State-of-the-art machine learning for real-world robotics in pytorch.
\newblock \url{https://github.com/huggingface/lerobot}, 2024.

\end{thebibliography}

\newpage
\appendix
\section*{Appendix}
\section{Datasets}\label{ap:model_details}

\subsection{Dataset for \model SSL Pretraining}

\begin{wrapfigure}{r}{5cm}
    \centering
    \includegraphics[width=\linewidth]{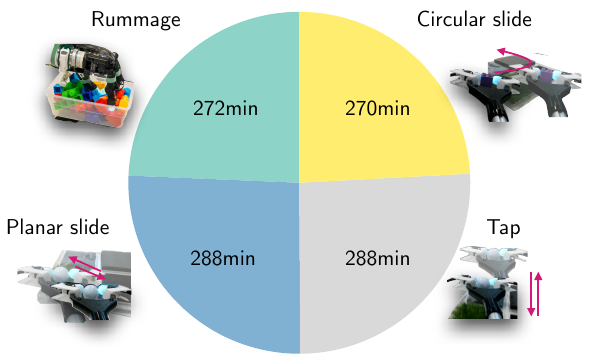}
    \caption{Distribution of recorded Digit~360 data by platform. The dataset includes 18.6 hours of data collected using two platforms: the Allegro hand (4.5 hours) and the mobile picker tool (14.1 hours).}
    \label{fig:dist_ssl}
\end{wrapfigure}

Our self-supervised learning (SSL) dataset is sourced from two platforms: an Allegro hand equipped with Digit~360 sensors mounted on each fingertip, and a custom mobile picker tool with sensors integrated into its gripping mechanism. Since SSL representation learning does not require labeled data, we collect tactile data by having the Allegro hand interact rummaging freely with a tray filled with LEGO blocks and marbles. This setup enables the capture of rich, multi-contact interactions with objects that feature distinctive geometries, such as spherical shapes and sharp edges.

The mobile picker is used to gather tactile data from everyday manipulation actions, including tapping and sliding, across surfaces with varying friction and stiffness. This allows us to record both intrinsic and extrinsic contact interactions. We collected eight sequences with the Allegro hand, each lasting approximately 8.5 minutes, recording data from all four fingers. From the mobile picker, we gathered 104 sequences with an average duration of 3 minutes, logging data from both sensors on the gripper. For the subset of the dataset collected with the mobile picker, we provide annotations indicating the object in grasp, the action performed, and the surface in contact, to support downstream evaluation tasks. In total (see Figure~\ref{fig:dist_ssl}), our dataset spans 18.6 hours of tactile data collected from six different Digit 360 sensors.

\begin{figure}
    \centering
    \includegraphics[width=\linewidth]{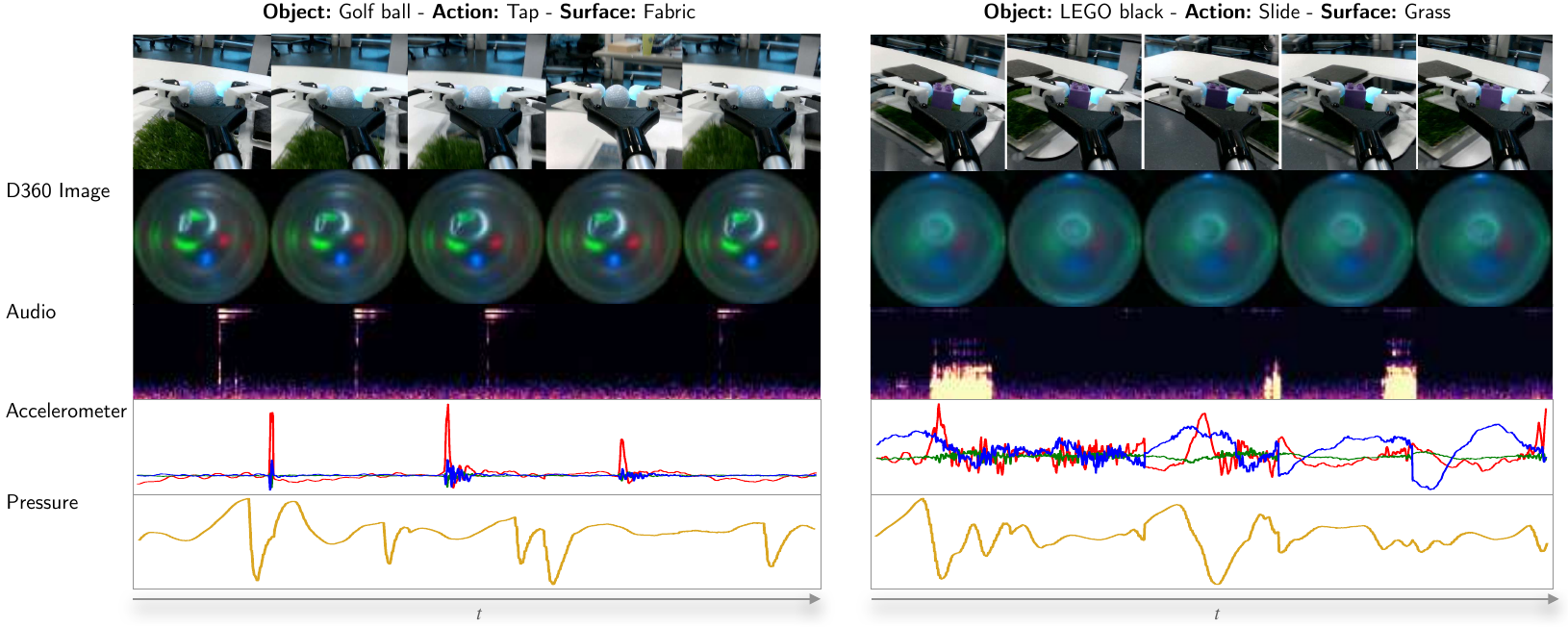}
    \caption{Visualization of each of the tactile input modalities to \model. Samples from pretraining dataset.}
    \label{fig:ssl_data_viz}
\end{figure}

\model processes temporal windows of data from each tactile modality. A visualization of the input data is shown in Figure~\ref{fig:ssl_data_viz}.

\textit{Images.} We input pairs of tactile images sampled with a temporal stride of 5, concatenated along the channel dimension. These images are captured using a hyperfisheye lens in Digit 360, allowing us to view the entire dome-shaped elastomer surface. Unlike planar GelSight-like sensors, these images include reflections from the surrounding LED light sources, visible near the center of the dome. While these reflections act as useful markers, encoding meaningful information about gel deformation upon contact, they also pose challenges for standard preprocessing techniques such as background subtraction or lighting augmentations, which risk corrupting the contact signal.

\textit{Audio.}  Digit~360 sensor is equipped with two contact microphones that capture vibrations, sampled at 48kHz. This signal is especially informative for detecting changes in contact state, such as making and breaking contact. \model processes 0.5sec windows of audio data from each microphone in the frequency domain. After standardization and conversion to log-mel spectrograms, the audio is treated as a single-channel image input to the model.

\textit{IMU and Pressure.} We extract 0.5 second and 1 second windows of data from the 3-axis accelerometer and the static pressure sensor embedded in the Digit~360. Each window is standardized using the mean and standard deviation computed per sequence to ensure consistency across variations in sensor signal amplitude.

\subsection{Datasets for Downstream Tasks}\label{ap:downstream_datasets}

For each experiment related to estimating physical properties with \model (see Section~\ref{sec:sparsh_d360_sl_tasks}), we designed custom setups to collect training data tailored to each task.

For \textbf{Object-Action-Surface Classification}, we repurpose the SSL pretraining dataset collected with the manual picker. We annotate each data point with metadata specifying the object being held (golf ball, wood block, LEGO block), the action performed (tap, linear slide, circular slide), and the external surface in contact (grass, fabric, plastic, foamwork). From the 104 available sequences, 69 are used for training and the remaining 35 for testing the performance of the classifier. The dataset is balanced in number of samples per label. 

\begin{figure}
    \centering
    \includegraphics[width=\linewidth]{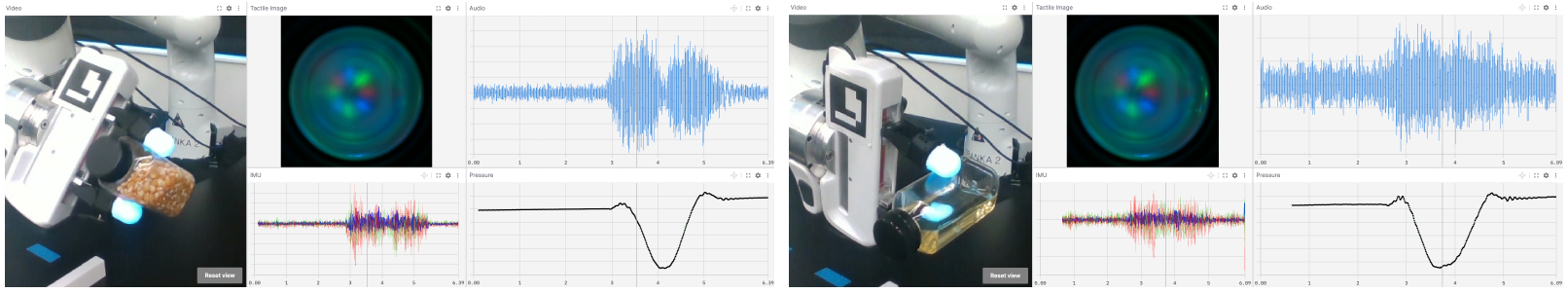}
    \caption{Visualization of the experimental setup and tactile sensory inputs for the material-quantity classification dataset. The setup involves shaking bottles filled with different materials and quantities using the Franka's gripper equipped with Digit~360 sensors. }
    \label{fig:material-quantity-data-illustration}
\end{figure}

\begin{figure}
    \centering
    \includegraphics[width=0.8\linewidth]{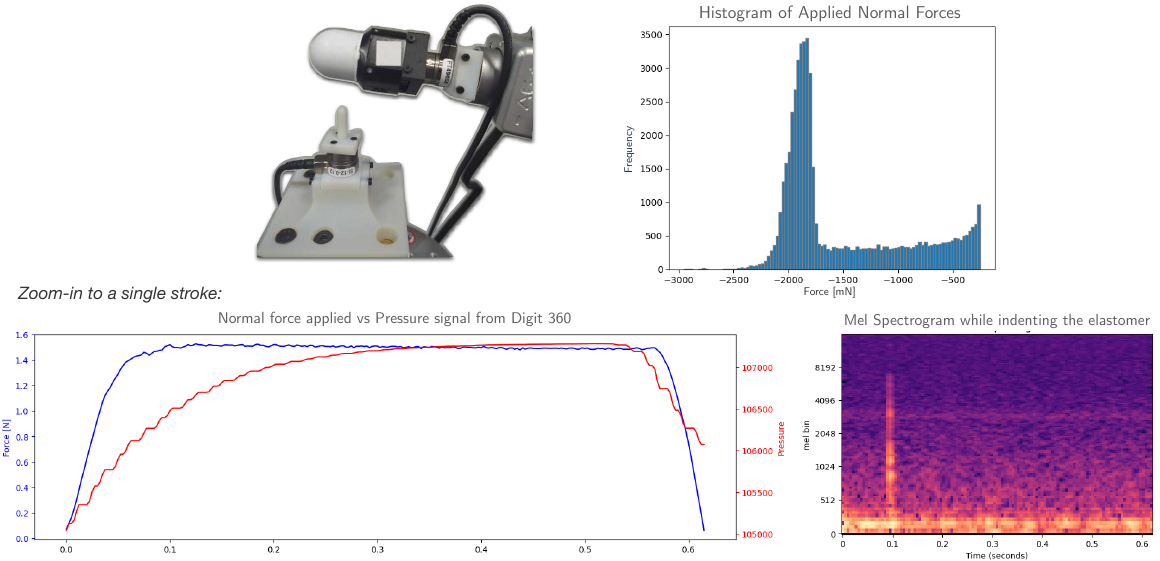}
    \caption{\textbf{Top:} Experimental setup and data distribution for the normal force regression experiment. \textbf{Bottom:} Zoom-in on a single indentation stroke. Note that the pressure signal from the Digit~360 sensor correlates well with the ground-truth normal force measured by the force/torque sensor beneath the hemispherical probe. The mel spectrogram also reveals the moment of contact between the probe and the elastomer.}
    \label{fig:force-data-illustration}
\end{figure}

For \textbf{Material-Quantity Estimation}, we design a 3D-printed gripper attachment to mount the Digit~360 sensors onto the Franka arm. The data collection protocol involves shaking six different 8oz bottles containing various materials (lentils, rice, corn kernels, vitamin pills, water, and oil) at different fill levels (full, half, quarter). An illustration of this setup is provided in Figure~\ref{fig:material-quantity-data-illustration}. For shaking the bottles to create variation in the tactile signal, the Franka’s gripper is rotated left and right in randomized motion patterns, including variation in the initial angle. We collect 20 trajectories for each material-quantity combination, using 15 sequences for training and reserving the remaining 5 for evaluating the classifier.

For \textbf{Normal Force Estimation}, we fix a hemispherical probe to a force/torque sensor and mount a Digit~360 sensor on the Meca arm, which is used to indent the elastomer surface perpendicularly, applying controlled normal forces of up to 3.5N. Figure~\ref{fig:force-data-illustration} illustrates the experimental setup and the distribution of the collected data. We observe that the pressure modality correlates strongly with both the magnitude of the applied normal force and the location of the resulting deformation on the elastomer. The audio modality captures discrete events, such as the initial contact between the probe and the sensor surface.

\section{Benchmarking \model for physical properties comprehension}

\begin{figure}
    \centering
    \includegraphics[width=\linewidth]{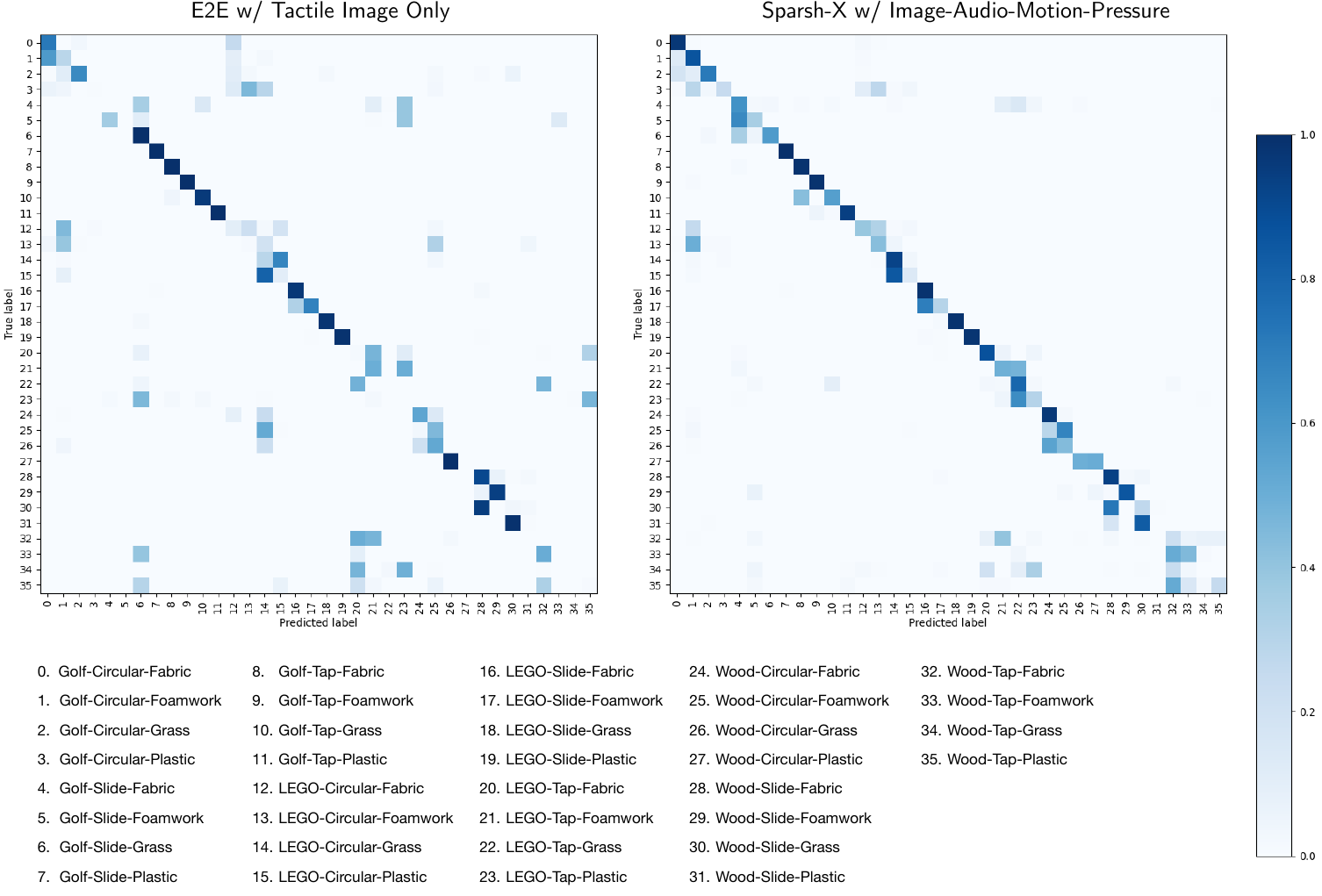}
    \caption{Confusion matrix for object-action-surface classification. We compare an end-to-end classifier trained solely on tactile images with a classifier trained on frozen \model representations, under a 50\% training data budget.}
    \label{fig:cm_obj_act_surf_clf}
\end{figure}

\paragraph{Object-Action-Surface Classification.} This task evaluates whether \model can capture tactile cues that enable the identification of objects through both intrinsic and extrinsic contact interactions. The goal is to jointly classify the object being grasped, the action performed, and the surface in contact. The selected objects and surfaces span a range of properties, including texture, hardness, and friction.

We use representations from \model to train a downstream classifier on the dataset described in Appendix~\ref{ap:downstream_datasets}. Figure~\ref{fig:cm_obj_act_surf_clf} shows confusion matrices on the test set for two classifiers: one trained using frozen \model representations with all tactile modalities as input, and another trained end-to-end~(E2E) using only tactile images. Note that the classification task involves 36 classes, representing all combinations of object, action, and surface. The results shown in the figure correspond to training with 50\% of the labeled training set.

The \model-based classifier shows stronger diagonal alignment, indicating more accurate predictions across the 36 joint object-action-surface classes. In contrast, the E2E model suffers from greater confusion among similar classes, particularly those with overlapping surface or action components (e.g., misclassifying "Tap-Foamwork" as "Tap-Fabric" or "Slide-Plastic"). These results highlight the benefit of multimodal tactile representations: incorporating audio, motion (IMU), and pressure modalities helps disambiguate fine-grained contact dynamics that are challenging to capture with images alone.

\paragraph{Material-Quantity Estimation. } This task further evaluates \model’s ability to comprehend physical properties. Specifically, we focus on distinguishing materials based on their granularity and viscosity (e.g., solids and liquids), as well as estimating mass through coarse volume classification. We train a classifier to predict one of 18 joint classes, each representing a unique combination of material type and quantity level. The classifier is trained either using frozen \model representations or end-to-end (E2E) from tactile images alone.

\begin{figure}
    \centering
    \includegraphics[width=\linewidth]{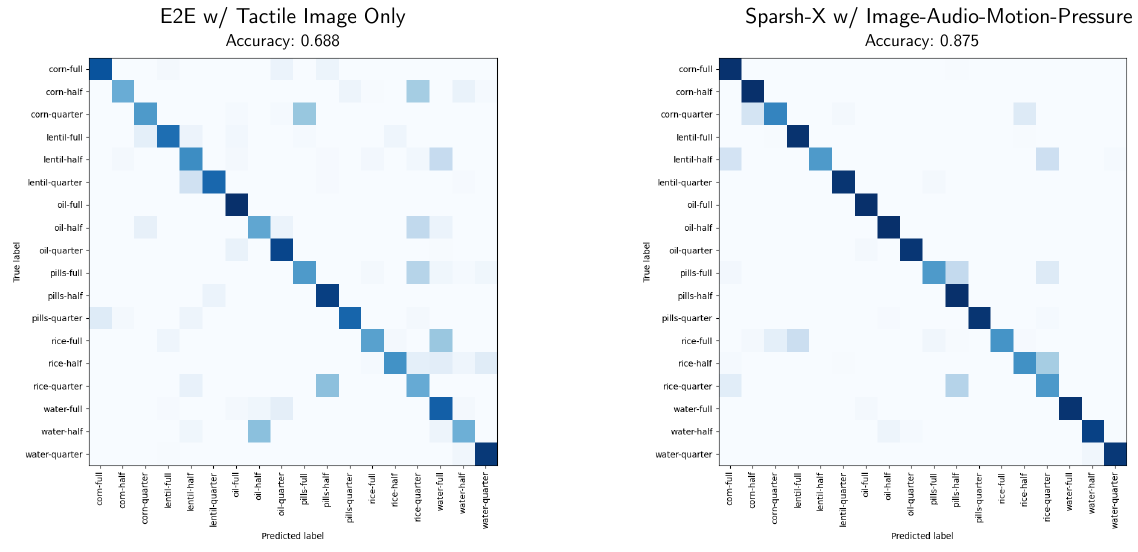}
    \caption{Confusion matrix for material-quantity estimation. We compare an end-to-end classifier trained solely on tactile images with a classifier trained on frozen \model representations, under a 33\% training data budget.}
    \label{fig:cm_grains_clf}
\end{figure}

Figure~\ref{fig:cm_grains_clf} shows the confusion matrices for the material-quantity classification task when trained with 33\% of labeled data, comparing an end-to-end (E2E) classifier trained solely on tactile images with a classifier trained on frozen \model representations. The E2E model achieves 68.8\% accuracy, while the \model-based classifier reaches 87.5\%, highlighting the benefit of multimodal tactile representations. In the E2E setting, we observe frequent confusion between different fill levels of the same material and between visually similar liquids such as oil and water. In contrast, the TacX-based classifier exhibits strong diagonal alignment, suggesting accurate identification across the 18 material-quantity classes.

\section{\model and Policy Learning}

\subsection{Real-World \model and Policy Deployment}
\begin{figure}
    \centering
    \includegraphics[width=0.8\linewidth]{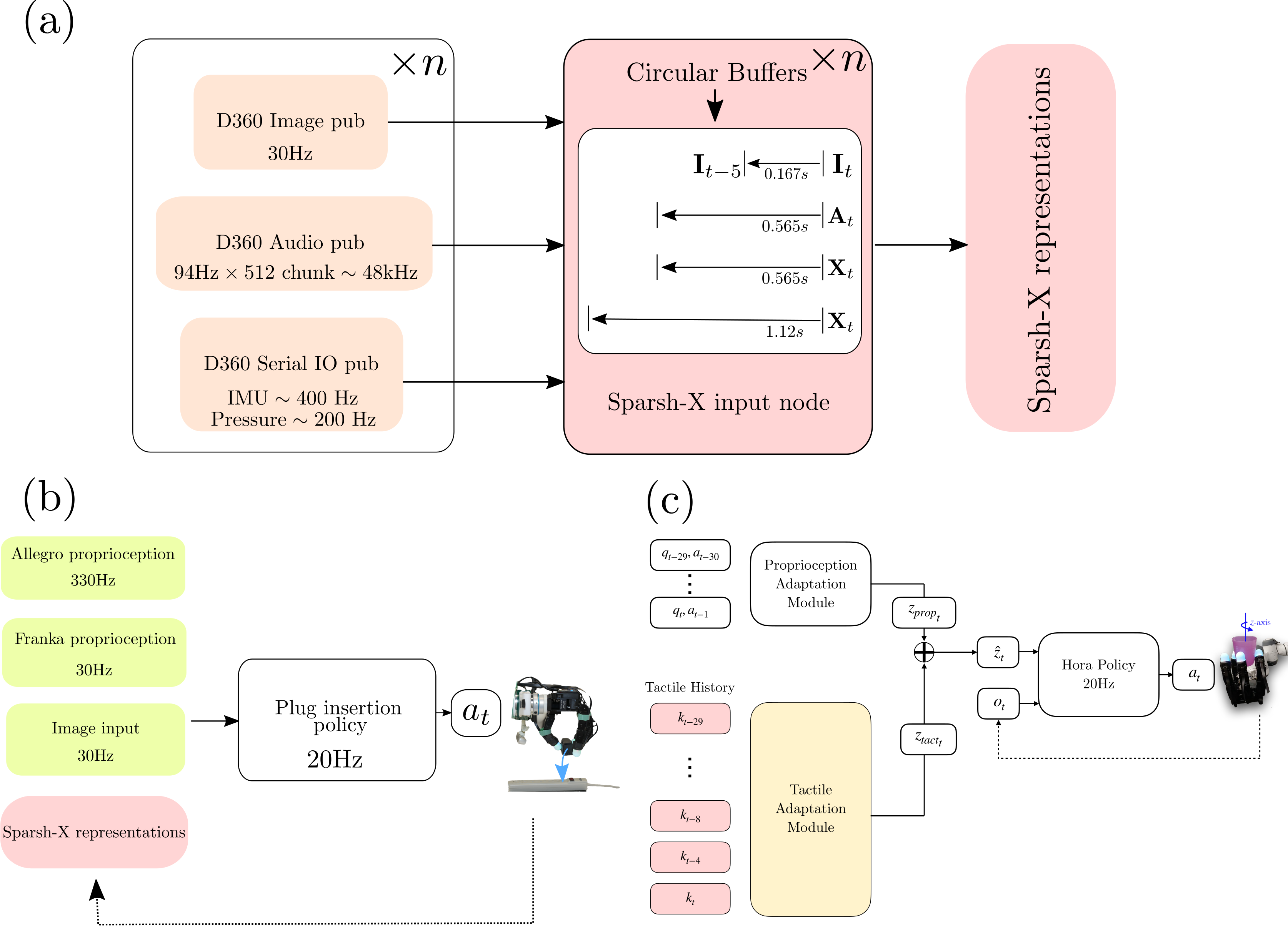}
    \caption{Real-world policy deployment architecture: We use ROS2 middleware for policy deployment, and PyTorch for deep learning modules. In addition to the proprioceptive states of the robot and optional third-person vision modality, downstream policies take as input \model representations from upto 4 fingertips of the Allegro hand. (a) illustrates how the inputs are constructed for TacX, (b) illustrates policy deployment for the plug insertion policy, and (c) illustrates the policy deployment for the in-hand rotation (Hora) policy. }
    \label{fig:model_deployment}
\end{figure}

For real-world deployment of \model, we use ROS2. We maintain circular buffers of 5 seconds for each tactile modality per Digit~360 fingertip. For synchronization, we use the timestamp of the image modality as the reference, selecting the closest-in-time sample from the other modalities (audio, motion from accelerometer, and pressure). Once inputs are processed, \model can run inference at 50Hz on a GPU RTX~4090. However, the construction of log-mel spectrograms remains the main computational bottleneck for real-time processing. When processing all four Digit~360 sensors on the Allegro hand, end-to-end inference with \model runs at approximately 20Hz. Figure~\ref{fig:model_deployment} shows the deployment pipeline for policy experiments.

\subsection{Plug-Insertion via Imitation Learning}

\paragraph{Training details.}  The robot, equipped with an Allegro hand and sensorized with Digit~360 fingertips, is tasked with inserting a pre-grasped plug into the first socket of an extension power strip. In our experimental setup, the socket position remains fixed, while the starting position of the robot arm is randomized within a 3D cuboid of $(5, 5, 2)~cm$ around the nominal starting pose.

\begin{figure}
    \centering
    \includegraphics[width=0.5\linewidth]{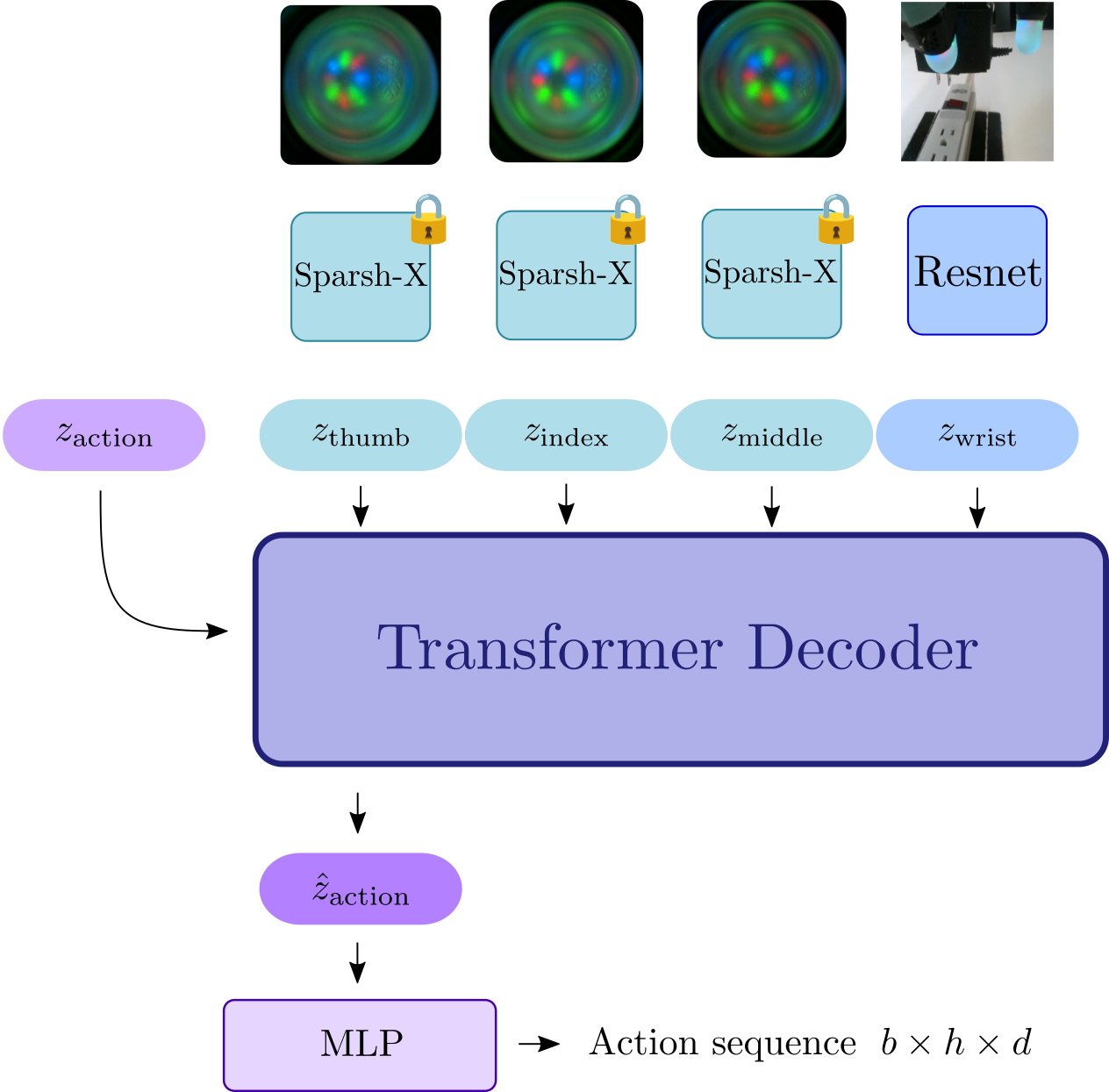}
    \caption{Architecture overview of the plug insertion policy. A transformer decoder is trained to generate action sequences based on \model representations from three fingertips, a wrist camera image capturing the current robot state, and a learnable latent action code. All embeddings are concatenated and processed by a lightweight MLP to decode the next end-effector action.}
    \label{fig:plug_insertion_arch}
\end{figure}

The model inputs include an embedding of the wrist camera image and \model representations for thumb, index, and middle finger sensors, processed through an attentive pooling layer~\cite{chen2024context}. We train a ResNet18~\cite{he2016deep} in an end-to-end (E2E) fashion to learn the wrist image embeddings. We append learnable action tokens, which are processed by the transformer and subsequently decoded into a sequence of actions. In our setup, the model predicts a sequence of absolute robot end-effector poses i.e., $\av \triangleq (\Tv_t, \Tv_{t+1}, \dots \Tv_{t+H})$ with a prediction horizon of $H=8$. An illustration of the plug insertion architecture is shown in Figure~\ref{fig:plug_insertion_arch}. 

\subsection{In-Hand rotation with sim-to-real tactile adaption}

\paragraph{Training details. }Hora~\cite{qi2022hand} is a two-stage policy that rotates objects along the \textit{z}-axis. The first stage trains the policy using privileged information, which includes the object's state or pose, local shape, mass, friction, and other physical properties that can be perceived by the fingertips. The second stage trains an adaptation module to approximate the latent space of the privileged information from the discrepancy between observed proprioception history and commanded actions, which implicitly informs about contact.

Although the approximation of the privileged vector from proprioceptions transfers to the real setup, it operates with incomplete information about the object's state. With multisensory touch sensing at the fingertip level, privileged information such as changes in object pose, slip, and friction are now accessible in real-world scenarios, albeit not directly.  We can leverage \model representations to fine-tune the real-world approximation of the privileged information embedding. The goal is to do \emph{tactile adaptation} on top of the baseline policy to enhance stability during object rotation.

We pass to the tactile adaptation module frozen \model representations for each of the four fingers in the Allegro-hand, with a temporal stride of 0.19s, equating to 8 touch representations per finger over a 1.5s window, which matches the proprioception state history consumed by the baseline Hora. Features for each finger are pooled using attentive pooling to create a global representation, which is then concatenated along the temporal dimension, resulting in a $(t\times n)\times 768$ input embeddings. The tactile adaptation model to be trained is a shallow MLP followed by the zero-convolution layer.

Our dataset consists of successful rollouts of the Hora policy, where the object keeps rotating without touching the palm for at least 30 seconds. The data is serialized into the lerobot dataset format~\cite{cadene2024lerobot}, sampled at a control frequency of 20Hz. For training the tactile adaptation module, our objective is to minimize the L2 loss between the real-world hand joint angles and the target action given by the frozen Hora policy under the tactile-informed privileged embedding.

\end{document}